\documentclass[letterpaper]{article} 
\usepackage{aaai2026}  

\nocopyright 
\usepackage{times}  
\usepackage{helvet}  
\usepackage{courier}  
\usepackage[hyphens]{url}  
\usepackage{graphicx} 
\usepackage{natbib}  
\usepackage{caption} 
\usepackage{algorithm}
\usepackage{algorithmic}
\usepackage{xcolor}
\usepackage{amsmath}
\usepackage{booktabs}
\usepackage{subcaption}
\usepackage{multirow}
\usepackage{amssymb}
\usepackage{afterpage}
\usepackage{newfloat}
\usepackage{listings}
\DeclareCaptionStyle{ruled}{labelfont=normalfont,labelsep=colon,strut=off} 
\floatstyle{ruled}
\newfloat{listing}{tb}{lst}{}
\floatname{listing}{Listing}
\title{3D Optimization for AI Inference Scaling: Balancing Accuracy, Cost, and Latency}

\author{Anonymous Authors}

\author{
    Minseok Jung\textsuperscript{\rm 1}\thanks{Contact: minseok.jung@cloudera.com.},   
    Abhas Ricky\textsuperscript{\rm 1},
    Rameez Chatni\textsuperscript{\rm 1}
}

\affiliations{
    \textsuperscript{\rm 1} Cloudera\\

    5470 Great America Parkway\\
    Santa Clara, CA 95054 USA\\

}

\usepackage{bibentry}

\begin{document}

\maketitle

\begin{abstract}
AI inference scaling is often tuned through 1D heuristics (a fixed reasoning pass) or 2D bivariate trade-offs (e.g., accuracy vs.\ compute), which fail to consider cost and latency constraints. We introduce a 3D optimization framework that jointly calibrates accuracy, cost, and latency within a unified decision space, enabling constraints-aware inference scaling. Using Monte Carlo simulations across three representative scenarios and nine simulated large language models, we evaluate four optimization methods to address the 3D multi-objective optimization (MOO) problem. Framing inference scaling in MOO shapes a feasible space that 1D and 2D optimizations fail to capture, enabling environment-adaptive selection of the inference scaling~$k$. Results show that knee-point optimization based on Pareto frontiers achieves the best balance, while accuracy-maximization remains favorable when accuracy is prioritized. Our results further show that smaller models, when combined with optimal inference scaling, can match or exceed the performance of larger models at a fraction of the cost. The framework establishes a theoretical foundation for deployment-aware inference scaling across diverse operational conditions. Code and simulation available at \url{https://github.com/masonjung/inference-scaling-moo}.
\end{abstract}


\section{Introduction}

Beyond scaling AI training, which enhances performance by providing more data, \emph{scaling inference}---which executes multiple reasoning paths before generating a final output---has been highlighted as an efficient method to improve performance~\citep{snell2024scaling} and reduce erroneous generations~\citep{park2025know}. By sampling multiple responses, instead of relying on a single generation, inference scaling increases the likelihood of obtaining higher-quality outputs at the inference stage. This enables even smaller models to surpass the larger models~\citep{wu2024inference,wang2022self}. This approach is particularly efficient when parallel computing is available, where advances in AI infrastructure allow multiple inference instances to run concurrently~\citep{bian2025scaling,nvidia2025balancing}.

\begin{figure}[t]
  \centering
  \includegraphics[width=\linewidth]{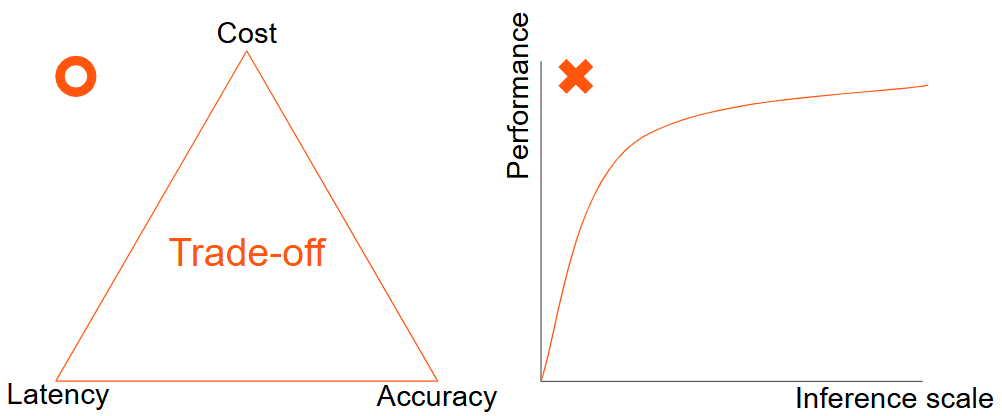}
  \caption{AI inference scaling requires optimizing the multidimensional tradeoff among cost, latency, and accuracy rather than relying on a bivariate performance-compute tradeoff that neglects the cost and latency constraints inherent in real-world deployment.} 
  \label{fig:3d_vs_2d}
\end{figure}

\begin{figure*}[t]
  \centering
  \includegraphics[width=0.88\linewidth]{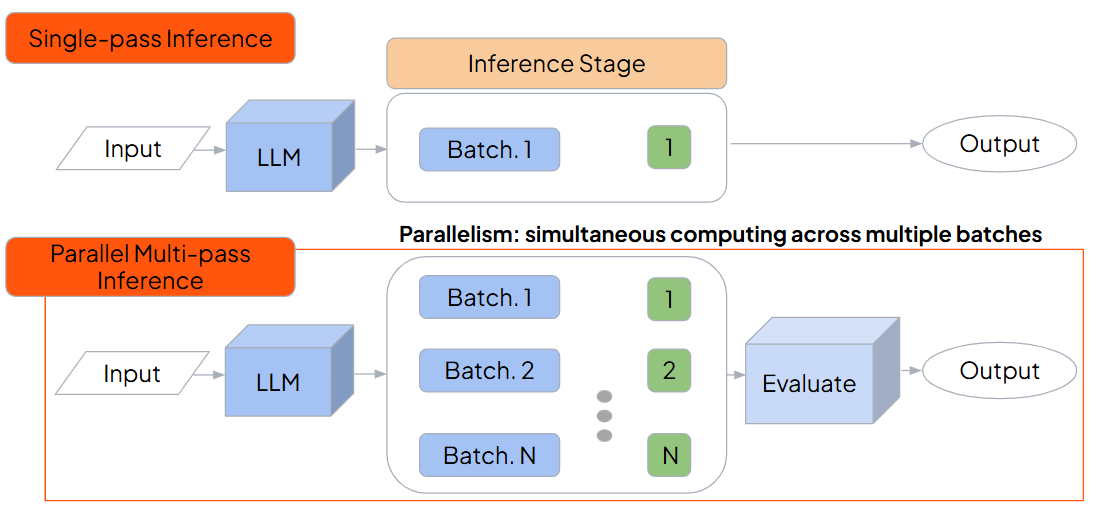}
    \caption{Unlike conventional inference, which performs a single stochastic forward pass (\(k{=}1\)) to generate one output, \emph{parallel inference scaling} executes multiple inference batches concurrently (\(k \ge 2\)) and aggregates their outputs (e.g., best-of-$k$) to improve accuracy with minimal increase in latency. Parallelism is particularly effective when sufficient computational resources are available to support concurrent execution.}
  \label{fig:parallelism}
\end{figure*}

Prior research on AI inference scaling has primarily focused on 2D relationships between performance and inference scale, demonstrating accuracy gains with increased computational allocation \cite{wang2022self, jin2025energy}. This approach addresses the limitation of static inference scaling by allowing adaptive compute allocation. For example, in medical AI applications, users may willingly invest additional computational resources to obtain more reliable results. Executing multiple inference passes substantially increases the likelihood of generating higher-quality outputs compared to the single-shot inference~\cite{snell2024scaling}. Such bivariate optimization effectively identifies the inference scale that maximizes accuracy under a limited compute budget. However, the 2D optimization fails to account for constraints on time and cost, factors that critically considered in the real-world deployment settings \cite{nvidia2025balancing}. Neglecting these dimensions renders existing scaling analyses incomplete for practical, deployment-aware optimization of AI inference.

We argue that inference scaling should be formulated as a \textbf{multidimensional optimization problem} (MOO) that jointly considers accuracy, cost, and latency, as illustrated in Fig.~\ref{fig:3d_vs_2d}. Conventional 2D overlooks the associated costs and latency constraints inherent in real-world deployments. For example, in clinical decision support systems, where inference must occur within strict latency and cost budgets, the model can perform only a limited number of reasoning passes, even if additional computation could marginally improve accuracy. To capture these effects more realistically, it is better to model the tradeoff in 3D space to be accountable to core features: \emph{performance, cost, and latency.}

To address the AI inference scaling problem from a multidimensional perspective, we develop a framework that integrates MOO, simulation for LLM generations, and Monte Carlo (MC) estimation. Building on the 2D bivariate tradeoff between performance-compute, we extend the notion of inference scaling into a tri-objective regime that jointly optimizes accuracy, cost, and latency. In the simulation, we model inference for a stochastic generation with parallelism, and formulate the search for the optimal scale $k^\star$ as a constrained MOO. We further run four optimization methods--accuracy maximization, cube-volume balance, and Pareto-based utopia closet and knee point selection--to characterize efficiency trade-offs under diverse deployment constraints.

The proposed framework provides a unified lens for understanding deployment-aware inference scaling. The main contributions of this research are as follows:
\begin{itemize}
    \item \textbf{MOO framework for inference scaling.} We formalize AI inference scaling as a multi-objective optimization (MOO) problem that jointly considers accuracy, cost, and latency. This framework explicitly incorporates cost and latency--factors overlooked in prior 1D and 2D optimizations--and establishes a theoretical foundation for deployment-aware inference optimization.

    \item \textbf{MC simulation under realistic constraints.} We develop a stochastic simulation that models LLM inference under realistic variability in token length, cost, and latency. Using MC estimation across nine representative model configurations and three constraint scenarios, we map optimality in the 3D objective space.
    
    \item \textbf{Comparative analysis of optimization strategies.} We evaluate four optimization methods—ACC maximization, maximal-cube, utopia-closest, and knee-point—simulating how each performs under different deployment priorities. The results show that knee-point selection achieves the best relative efficiency, while ACC- maximization is the best when accuracy is prioritized.
\end{itemize}

This MOO perspective to AI inference scaling not only overcomes the limitation of the bivariate tradeoff optimization but also bridges theoretical scaling-law and real-world deployment needs.

\section{Related Work}
Research on AI inference scaling and optimization has progressed along three major directions:  
(1) scaling inference through parallelism,  
(2) addressing operational constraints in deployment environments, and  
(3) formalizing inference selection as a MOO problem.  
These research threads are deeply interrelated: scaling techniques enable performance gains, system constraints delineate the feasible optimization space, and MOO provides principled methods for balancing the competing objectives.

\paragraph{Inference Scaling through Parallelism.}
Inference scaling methods enhance model performance by generating multiple candidate outputs and selecting the best among them. Techniques such as best-of-$k$ and self-consistency demonstrate that additional sampling at inference time can enhance reasoning capacity without retraining~\cite{sadhukhan2025kinetics,wang2022self}. Parallelism makes this attainable by executing multiple inference passes simultaneously with minimal latency overhead, leveraging distributed compute resources such as GPUs~\cite{nvidia2025balancing}. Recent studies also suggest that smaller models can match or even exceed the accuracy of larger ones when parallel inference is effectively utilized~\cite{snell2024scaling}. While these techniques highlight the potential of scaling inference through parallelism, most studies focus on increasing performance via more computing, rather than configuring the optimality with both cost and latency.

\paragraph{Operational Constraints and System Efficiency}
Another research line emphasized the need to incorporate cost and latency factors for real-world AI systems~\cite{nvidia2025balancing,aubrey2025maximizeai}. These studies highlight that inference performance is bounded by operational factors such as per-query cost ceilings and service-level latency requirements. Under such constraints, balancing performance, latency, and cost is crucial to achieve effective AI infrastructure operation ~\cite{harris2025thinksmart}. Most of the research remains grounded in bivariate performance-compute tradeoff curve~\cite{brown2024large, snell2024scaling,wang2022self, jin2025energy, park2025know}. To the best of our knowledge, no prior research has formulated inference optimization that simultaneously accounts for cost, latency, and performance, despite increasing recognition of its necessity.


\paragraph{MOO for Tri-Objective Inference Scaling.}
From an optimization standpoint, MOO provides an essential framework for balancing multiple competing factors in AI inference. Classical MOO approaches—including distance-to-utopia selection, hypervolume maximization, and knee-point detection—identify Pareto-optimal configurations that maintain balanced efficiency under resource constraints. Also, accuracy-maximization serves as a useful upper bound when performance dominates other considerations, whereas Pareto-based criteria better capture realistic deployment trade-offs. Building on these developments, our work explicitly formulates accuracy, cost, and latency as interdependent objectives within a unified MOO framework, enabling principled and deployment-aware determination of the optimal inference scale~$k$.

\begin{figure}[t]
  \centering
  \includegraphics[width=\linewidth]{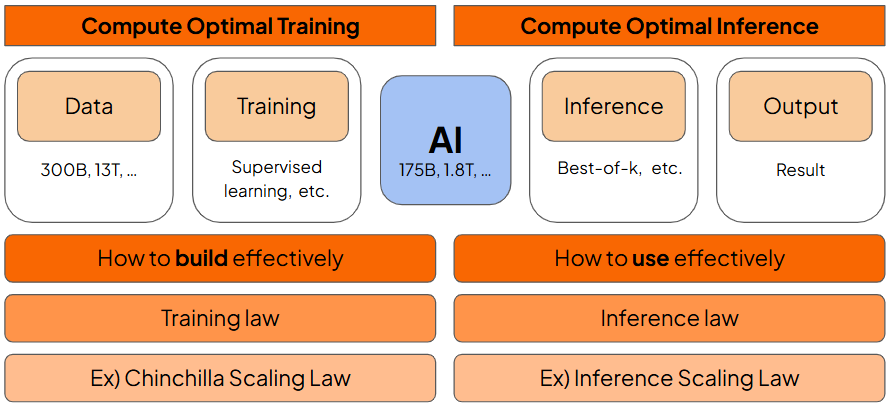}
  \caption{Compute-optimal AI system requires scaling efficiency in both training and inference. Training scaling laws (e.g., Chinchilla law) guide how models are built, while inference scaling laws govern how they are used efficiently under real-world cost and latency constraints.}
  \label{fig:train_inference}
\end{figure}

\section{Problem Formulation and Modeling}

\subsection{Overview}
The objective is to find the optimal inference scale $k$ that jointly balance \emph{accuracy}, \emph{cost}, and \emph{latency} under resource constraint scenarios. We formalize this as an MOO problem with multiple constraints. The key parameters are summarized in Table~\ref{tab:notation}.

\begin{table}[h]
\centering
\begin{tabular}{ll}
\toprule
\textbf{Symbol} & \textbf{Definition} \\
\midrule
$k$ & Number of inference passes \\
$L_{\text{in}},L_{\text{out}}$ & Input/output token lengths \\
$\mu_{L_{\text{in/out}}},\sigma_{L_{\text{in/out}}}$ & Mean and std.\ of in/out token lengths \\
$A_i$ & ACC of the $i$-th inference \\
$A(k)$ & Aggregate ACC from $k$ (e.g. best-of-$k$) \\
$\mu_A,\sigma_A$ & Mean and std.\ of single-inference ACC \\
$C_i,T_i$ & Cost and time of the $i$-th inference \\
$c_{\text{in/out}}$ & Cost coefficients for in/out tokens \\
$t_{\text{in/out}}$ & Latency coefficients for in/out tokens \\
$P$ & Parallel factor (concurrent inferences) \\
$C_{\max},T_{\max}$ & Maximum cost/time constraints \\
$A_{\min}$ & Minimum acceptable ACC\\
$\hat{\mu}_C(k)$ & Expected cost (MC) \\
$\hat{\mu}_T(k)$ & Expected time (MC) \\
$\hat{\mu}_A(k)$ & Expected ACC (MC) \\
$\mathcal{F}$ & Feasible set of inference counts \\
\bottomrule
\end{tabular}
\caption{Notation summary for the simulation. Symbols represent model-dependent quantities for ACC, cost, and latency. MC estimates are denoted by the hat symbol (\(\hat{h}\)).}
\label{tab:notation}
\end{table}

\subsection{Stochastic Model of Inference Scaling}
Each inference takes input tokens $L_{\text{in}}$ and generates output tokens $L_{\text{out}}$, drawn from Gaussian distributions:

\begin{align}
(L_{\text{in}}, L_{\text{out}}) \sim \big(\mathcal{N}(\mu_{L_{\text{in}}},\sigma_{L_{\text{in}}}^2), \mathcal{N}(\mu_{L_{\text{out}}},\sigma_{L_{\text{out}}}^2)\big)
\end{align}

with a clipping to ensure positivity. The accuracy of a single inference is also modeled as a Gaussian random variable:
\begin{equation}
A_i \sim \mathcal{N}(\mu_A,\sigma_A^2), \quad A_i\in[0,1].
\end{equation}

This assumption of parametric distribution is justified by the Central Limit Theorem (CLT) because both token lengths and single-inference ACC result from the numerous independent stochastic factors. The model is therefore appropriate for applying the CLT assumption, as these aggregated effects yield approximately normal variability around their means. When $k$ independent inferences are executed, aggregate performance is defined by an evaluation metrics. We use the ``best-of-$k$'' rule:
\begin{equation}
A(k)=\max\{A_1,A_2,\dots,A_k\}.
\end{equation}
The ``best-of-k'' is the most common AI inference strategy~\citep{snell2024scaling}. 

\subsection{Cost and Latency per Inference}
The costs and latency of an inference are modeled as linear functions. For the $i$-th inference, the cost and latency are:

\begin{align}
C_i = c_{\text{in}} L_{\text{in},i} + c_{\text{out}} L_{\text{out},i};\quad T_i = t_{\text{in}} L_{\text{in},i} + t_{\text{out}} L_{\text{out},i}
\end{align}

In parallel computing environment, the $k$ inferences are executed concurrently (e.g., eight reasoning batches simultaneously) as shown in Fig. \ref{fig:parallelism}. Thus the latency can be divided by the parallel factor $P$, while each inference cost is same as a sequential execution:
\begin{align}
C(k) &= \sum_{i=1}^{k} C_i, \\
T(k) &= \frac{1}{P}\sum_{i=1}^{k} T_i 
     = \left(\frac{k}{P}\right)\bar{T}, 
\quad 
\bar{T}=\frac{1}{k}\sum_{i=1}^{k} T_i.
\end{align}

In this equation, $\bar{T}$ denotes the mean per-inference latency, and $P$ represents the degree of parallelism, reflecting the assumption that $P$ independent inference processes can run concurrently on separate compute units.

\subsection{Monte Carlo Approximation}
Because $C(k),T(k),A(k)$ are stochastic, closed-form expectations are generally intractable. That is to say, their values fluctuate across inference runs due to variability in usage. We therefore employ MC estimation with $M$ trials to approximate the mean and standard deviation.
\vspace{-2mm}
\begin{equation}
\hat{\mu}_q(k)
= \frac{1}{M}\sum_{j=1}^{M} q^{(j)}(k),
\quad q\!\in\!\{C,T,A\}.
\end{equation}

Confidence intervals (CI) (e.g., 95\%) could be also estimated from percentiles across trials.

\subsection{Feasible Region and Objective}
Deployment settings typically impose an upper bound on computational cost ($C_{\max}$) and latency ($T_{\max}$), as well as a required minimum accuracy ($A_{\min}$). These constraints jointly define the feasible set of inference counts:
\vspace{1mm}
\begin{equation}
\mathcal{F}=\{k \mid 
\hat{\mu}_C(k)\le C_{\max},\ 
\hat{\mu}_T(k)\le T_{\max},\ 
\hat{\mu}_A(k)\ge A_{\min}\}
\end{equation}

Within this feasible region, the inference scaling problem seeks the optimal scale of inferences $k^\star$:
\begin{equation}
\begin{aligned}
\text{Given } & \big(\hat{\mu}_A(k),\hat{\mu}_C(k),\hat{\mu}_T(k)\big), \\
\text{Find } & k^\star \in \mathcal{F}
\end{aligned}
\end{equation}
such that $k^\star$ maximizes the joint objective $f(\hat{\mu}_A(k),\hat{\mu}_C(k),\hat{\mu}_T(k))$, where $f(\cdot)$ is defined by an optimization method introduced in Section~\ref{sec:method}.

\subsection{Scenario under Different Budgets}
To simulate the MOO under the realistic 3D constraints, we describe three representative budget scenarios in Table \ref{tab:scenario}.

\begin{table}[h]
\centering
\begin{tabular}{lccc}
\hline
\textbf{Scenario} & \textbf{Cost} & \textbf{Time} & \textbf{Accuracy} \\
\hline
1: Essay feedback & Low & Low & Moderate \\
2: Medical AI & High & High & Very high \\
3: Proposal writing & Low & High & High \\
\hline
\end{tabular}
\caption{Deployment scenarios showing how cost, time, and accuracy priorities differ across domains.}
\label{tab:scenario}
\end{table}

\subsubsection{Scenario 1: Low cost and latency (Essay feedback).}
In educational feedback applications, users prefer quick, low-cost responses over higher-priced and slower outputs. Since minor phrasing inconsistencies are acceptable, this scenario tightens the feasible region $\mathcal{F}$ by imposing stricter cost and time constraints.

\subsubsection{Scenario 2: High cost and time budget (Medical AI).} 
In high-stakes domains such as medical decision support, users are open to allocate greater budgets for both cost and time while very high accuracy is demanded. Here, accuracy maximization naturally pushes toward the highest feasible set $k$.

\subsubsection{Scenario 3: Low cost and high time budget (Proposal writing).}  
In proposal drafting, latency constraints are relaxed, allowing longer inference durations for improved accuracy or style, while cost remains limited due to the exploratory nature of the task. Thus, the feasible region $\mathcal{F}$ prioritizes low-cost efficiency and tolerates higher latency, making accuracy maximization more viable.

\vspace{2mm}
These scenarios illustrate that the choice of optimization method depends on the deployment context, as differing priorities for cost, speed, and performance necessitate distinct definitions of optimal inference scaling for each case.

\begin{figure*}[t]
  \centering
  \begin{subfigure}{0.32\textwidth}
    \includegraphics[width=\linewidth]{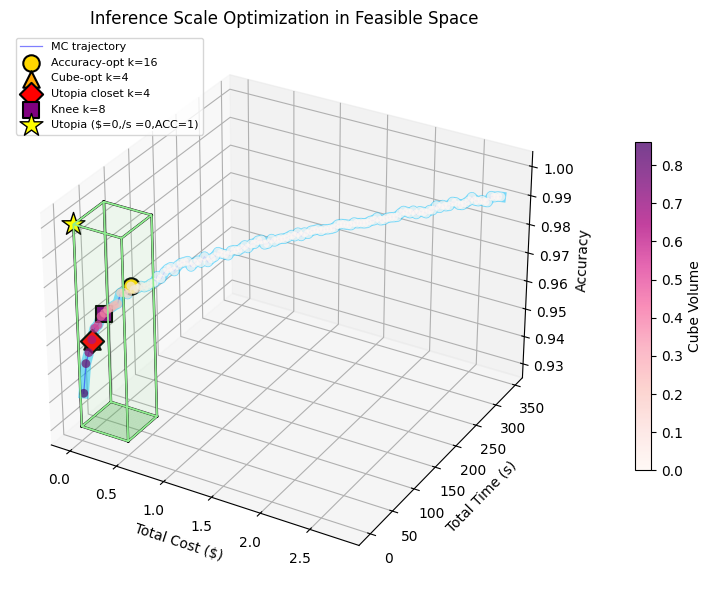}
    \caption{3D MOO results for Case~1.}
  \end{subfigure}
  \begin{subfigure}{0.32\textwidth}
    \includegraphics[width=\linewidth]{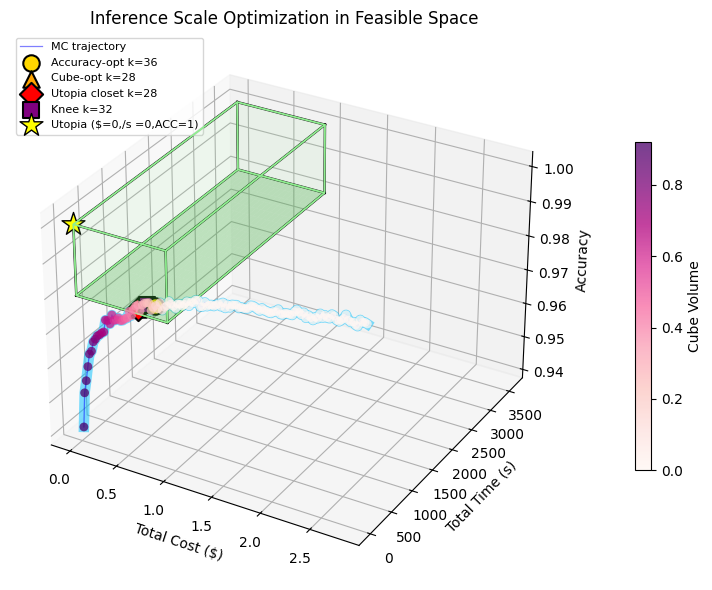}
    \caption{3D MOO results for Case~2.}
  \end{subfigure}
  \begin{subfigure}{0.32\textwidth}
    \includegraphics[width=\linewidth]{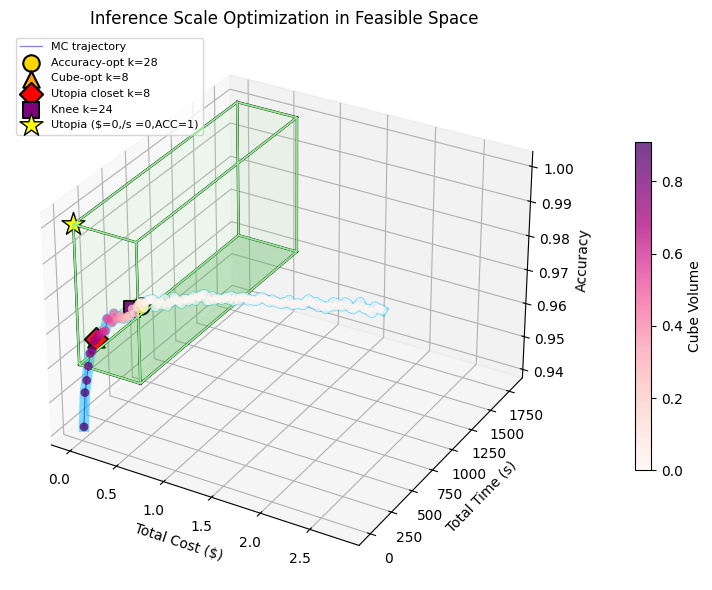}
    \caption{3D MOO results for Case~3.}
  \end{subfigure}
    \caption{
    Inference-scaling optimization results for GPT-5 across three simulated scenarios. 
    (a--c) 3D feasible cubes in cost--time--accuracy space with constraint planes of maximum cost, maximum latency, and minimal ACC (\(C_{\max}\), \(T_{\max}\), and \(A_{\min}\)). Markers denote each optimal point (\(\bullet\) Maximum-ACC, \(\blacktriangle\) Cube-Optimal, \(\blacklozenge\) Utopia-Closest, \(\blacksquare\) Knee-Point); purple colors indicate cube volume (larger is better), and the yellow star marks the utopia point.}
  \label{fig:3d_results}
\end{figure*}

\begin{figure*}[t]
  \centering
  \begin{subfigure}{0.32\textwidth}
    \includegraphics[width=\linewidth]{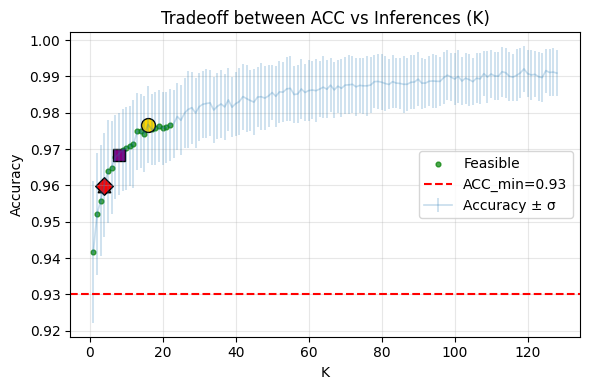}
    \caption{2D ACC--k with optimality for Case~1.}
  \end{subfigure}
  \begin{subfigure}{0.32\textwidth}
    \includegraphics[width=\linewidth]{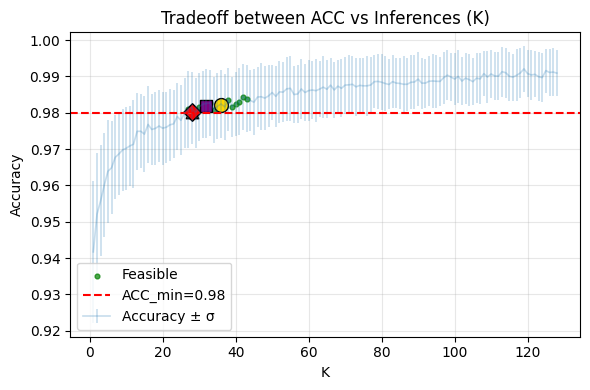}
    \caption{2D ACC--k with optimality for Case~2.}
  \end{subfigure}
  \begin{subfigure}{0.32\textwidth}
    \includegraphics[width=\linewidth]{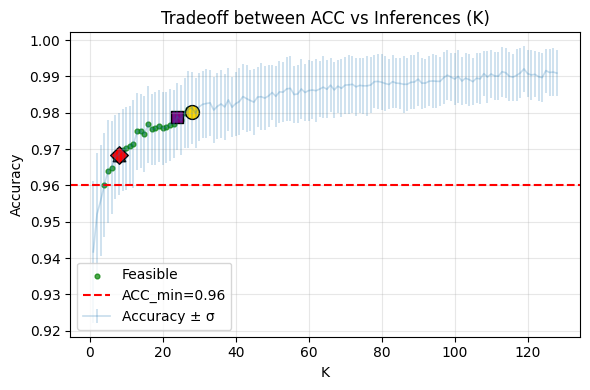}
    \caption{2D ACC--k with optimality for Case~3.}
  \end{subfigure}
    \caption{Accuracy-inference scale (k) trade-offs across three scenarios for GPT-5. These show MC means with CI = 95\%; the red dashed line indicates minimal ACC (\(A_{\min}\)). Markers correspond to the optimal configurations identified in Fig.~\ref{fig:3d_results}. Distinct operational priorities yield different optimal inference scales \(k^\star\).}
  \label{fig:gpt5_results}
\end{figure*}

\subsection{Example Models for Simulations}
We simulate nine model configurations derived from three sample LLM families—GPT-5, Nemotron, and Qwen3—each instantiated in large, medium, and small sizes. These models capture a broad spectrum of contemporary inference settings with varying performance, latency, and cost. This setup provides a unified basis for analyzing how different budget regimes (Table~\ref{tab:scenario}) influence the optimal inference scale~$k^\star$ under MOO.

\section{Methodology}\label{sec:method}

\subsection{Overview}
Given the estimated metrics $\big(\hat{\mu}_A(k),\hat{\mu}_C(k),\hat{\mu}_T(k)\big)$, 
we identify the optimal inference count $k^\star$ within the feasible set $\mathcal{F}$. 
We evaluate four optimization strategies:  
(1) \emph{Accuracy Maximization},  
(2) \emph{Maximum Cube Volume},  
(3) \emph{Utopia-Closest Selection}, and  
(4) \emph{Knee-Point Selection}.  
Each embodies a distinct trade-off measure in the 3D space.

\subsection{Accuracy Maximization}
The most direct approach maximizes expected accuracy:
\begin{equation}
k^\star=\arg\max_{k\in\mathcal{F}}\hat{\mu}_A(k).
\end{equation}
This represents an upper bound that maximizes the ACC while giving a negligible weight on cost and time. 

\subsection{Maximum Cube Volume}
To balance all three objectives in the maximal cube volume in the 3D space, we define normalized goodness scores:
\begin{equation}
g_A(k)=\hat{\mu}_A(k),\quad 
g_C(k)=1-\tfrac{\hat{\mu}_C(k)}{C_{\max}},\quad 
g_T(k)=1-\tfrac{\hat{\mu}_T(k)}{T_{\max}}.
\end{equation}

The cube-volume objective maximizes their product:
\begin{equation}
k^\star=\arg\max_{k\in\mathcal{F}} g_A(k)\,g_C(k)\,g_T(k),
\end{equation}
favoring balanced solutions that make the largest cube under the space rather than prioritizing one or two features.

\subsection{Pareto Frontier Optimizations}
We adopt a Pareto-frontier approach that identifies configurations where no objective can improve without degrading another, representing the best achievable balance among accuracy, cost, and latency. Formally, Pareto frontiers $p_2=(C_2,T_2,A_2)$ dominates others $p_1=(C_1,T_1,A_1)$ if \begin{equation}
C_2 \le C_1,\quad T_2 \le T_1,\quad A_2 \ge A_1,
\end{equation}
with at least one strict inequality. 
The Pareto frontier is then defined as
\begin{equation}
\mathcal{P} = \{\, p \in \mathcal{F} \mid \nexists\, q \in \mathcal{F} : q \succ p \,\}.
\end{equation}
$\mathcal{P}$ represents the boundary of optimal trade-offs where improving one metric inevitably worsens another, forming the efficient surface in the cost–time–accuracy.

\subsubsection{Utopia-Closest Selection}
The utopia point represents an unattainable ideal corresponding to perfect accuracy, zero cost and latency. We can get an efficient solution by picking up a point from the Pareto-frontiers that is the most close to the utopia point that is defined as
\begin{equation}
p^\star = (time = 0, cost = 0, accuracy = 1).
\end{equation}
It is ideal but unreachable. But this point serves as a geometric reference for evaluating Pareto-optimal points. Each Pareto point is normalized to a comparable scale as
\begin{equation}
\tilde{p} = \Big(\tfrac{C}{C_{\max}},\, \tfrac{T}{T_{\max}},\, 1 - A\Big),
\end{equation}
and the solution closest to this ideal is selected as
\begin{equation}
p_{\text{utopia}} = \arg\min_{p \in \mathcal{P}} \|\tilde{p} - p^\star\|.
\end{equation}
This distance-based criterion selects the configuration that lies nearest to the utopia point on the Pareto surface.

\subsubsection{Knee-Point Selection}
The knee point represents the region of maximum curvature on the Pareto surface. For example, the knee point marks the moment where increasing accuracy begins to yield diminishing returns relative to the rising cost and latency, representing the relative efficiency. Let $\mathbf{p}(k) = [\tfrac{\hat{\mu}_C(k)}{C_{\max}}, 
\tfrac{\hat{\mu}_T(k)}{T_{\max}}, 1-\hat{\mu}_A(k)]$ 
denote the normalized trajectory of inference scaling. The curvature is then:
\begin{equation}
\kappa(k) = 
\frac{\|\mathbf{p}'(k) \times \mathbf{p}''(k)\|}
{\|\mathbf{p}'(k)\|^3}.     
\end{equation}
Intuitively, $\mathbf{p}'(k)$ measures how the trade-off surface changes as $k$ increases, and $\mathbf{p}''(k)$ captures how rapidly that change bends. The denominator normalizes for scale, ensuring curvature reflects the shape of the trade-off rather than its magnitude. The knee point corresponds to:
\begin{equation}
k_{\text{knee}} = \arg\max_k \kappa(k),
\quad
p_{\text{knee}} = \mathbf{p}(k_{\text{knee}}).
\end{equation}
where $k_{\text{knee}}$ denotes the inference count at which curvature is maximal, and $p_{\text{knee}}$ represents the corresponding cost–time–accuracy point on the Pareto surface. The optimal knee point is found by locating the inference count $k_{\text{knee}}$ where curvature is maximal and then mapping it to its corresponding trade-off coordinates $p_{\text{knee}}$ on the Pareto surface.

\subsection{Rationales for Optimization Method Selection}
These four criteria capture complementary perspectives:  
\textbf{Accuracy Maximization} targets peak performance,  
\textbf{Cube Volume} enforces balanced efficiency,  
\textbf{Utopia-Closest} emphasizes geometric closeness to an ideal,
and \textbf{Knee-Point} locates the inflection where marginal accuracy gains demand disproportionate cost or time. Because deployment priorities vary (e.g., latency-sensitive, cost-limited, and flawlessness), no single criterion universally dominates. These optimization methods provide practical solutions.

\subsection{Simulation Overview and Hyperparameters}
All simulations adopt consistent parameterization for comparability. Input lengths follow $\mathcal{N}(1024, 64^2)$ and output lengths $\mathcal{N}(2048, 128^2)$; parallelism factor $P=4$; Monte Carlo trials $M=300$; random seed $42$; and maximum inference count $k_{\max}=128$. Each scenario applies distinct budget constraints on maximum cost ($C_{\max}$), maximum latency ($T_{\max}$), and minimum acceptable accuracy ($A_{\min}$). 
Full model configurations are listed in Appendix~\ref{tab:model_configs}.

\section{Results}\label{body_results}
\subsection{Simulation Overview}
We systematically evaluated four optimization criteria across three deployment constraint regimes and nine representative simulated LLMs drawn from the GPT-5, NVIDIA Nemotron, and Qwen3 families noted in Appendix~\ref{tab:scenario1}. 

\subsection{Results Across Deployment Scenarios}
Across scenarios, Knee-point optimization produced the most compute-efficient solutions under constrained settings (Scenarios~1 and~3), while accuracy-maximization is the most suite for Scenario~2.

Figures~\ref{fig:3d_results} and~\ref{fig:gpt5_results} visualize the corresponding Pareto surfaces in the accuracy–cost–latency space of GPT-5 and conventional 2D tradeoff analysis from Scenario 1. The outcomes are summarized below and in Appendix 2.

\paragraph{Scenario~1: Low Cost and Latency Budget (Essay Feedback).}
With $C_{\max}{=}0.50$, $T_{\max}{=}60$\,s, and $A_{\min}{=}0.93$, the accuracy-optimal configuration occurred at $k{=}16$ ($97.7\pm\!1.1\%$, \$0.349, $43.1$\,s). Cube-optimal and utopia-closest converged at $k{=}4$ ($96.0\!\pm\!1.4\%$, \$0.087, $10.8$\,s), whereas the knee-point was identified at $k{=}8$ ($96.8\!\pm\!1.2\%$, \$0.174, $21.5$\,s).  

\paragraph{Scenario~2: High Cost and Time Budget (Medical AI).}
For $C_{\max}{=}0.95$, $T_{\max}{=}3600$\,s, and $A_{\min}{=}0.98$, accuracy-maximization yielded the best absolute accuracy at $k{=}36$ ($98.2\!\pm\!0.9\%$, \$0.783, $96.7$\,s). 
Knee-point reached $k{=}32$ ($98.2\!\pm\!1.0\%$, \$0.696, $86.0$\,s) and cube-max and utopia-closet suggested $k{=}28$ ($98.0\!\pm\!1.0\%$, \$0.609, $75.2$\,s). 

\paragraph{Scenario~3: Low Cost, Flexible Latency (Proposal Writing).}
With $C_{\max}{=}0.65$, $T_{\max}{=}1800$\,s, and $A_{\min}{=}0.96$, the accuracy-optimal configuration occurred at $k{=}28$ ($98.0\!\pm\!1.0\%$, \$0.609, $75.2$\,s). The knee-point emerged at $k{=}24$ ($97.9\%\!\pm\!1.0\%$, \$0.522, $64.5$\,s). Cube- and utopia-closest selections ($k{=}8$) showing ACC of $0.968\!\pm\ 0.012\%$ with total cost of 0.174 in 21.5\,s.

\subsection{Results Across Models}
All model families exhibit similar Pareto curvature: larger architectures achieve higher baseline accuracy but incur steeper cost and latency growth as inference multiplicity increases. Smaller models attain comparable accuracy with lower compute budget, reflecting greater efficiency rather than higher absolute performance.

\paragraph{GPT-5 Family.}
For the GPT series, efficient compute allocation compensates for model size differences. GPT-5~Nano achieved $99.0\%$ accuracy at $k{=}56$ from the knee method exceeds GPT-5~Large’s $98\%$ at $k{=}16$ from Max-ACC while operating at approximately $1/7$ of the total cost (\$0.05 vs.\ \$0.35) under with lower latency (30.1/s vs. 43.1/s) in case 1.

\paragraph{NVIDIA Nemotron Family.}
Nemotron models exhibit a clear efficiency trend in Scenario~2, demonstrating that smaller architectures can match the accuracy of much larger ones when inference parallelism is leveraged. Nemotron~Nano~9B reached perfect accuracy at $k{=}52$ (from Max-ACC) with a total cost of \$0.117 and latency of $33.5$\,s earlier than the Ultra~253B model achieved the same accuracy at $k{=}72$ (from Max-ACC) with a cost of \$0.479 and latency of $386.9$\,s. This four-fold cost difference and ten-fold latency reduction underscore the efficiency of allocating compute to multiple inference paths in smaller models.

\paragraph{Qwen3 Family.}
Qwen3 models display stable and controllable scaling behavior. 
In Scenario~3, Qwen3-Max achieved perfect accuracy at $k{=}108$ (\$0.63, $464.6$\,s) with Max-ACC and $0.99$ at $k{=}56$ (\$0.33, $240.8$\,s), which is from Knee optimization, demonstrating efficiency of knee method in accuracy–cost-latency trade-offs, by halving cost and latency with a sacrifice of 1\% of ACC in scenario 3.

\subsection{Aggregate Trends- Knee}
Across all nine model configurations and three constraint regimes, knee-point optimization consistently produced the most cost-efficient trade-offs, reducing mean latency by $63\%$ and total cost by $58\%$ relative to accuracy-maximization, while maintaining accuracy within $1$–$2\%$. 
Cube-volume and utopia-closest methods offered additional savings in cost and latency with a tradeoff in ACC. These trends establish a consistent pattern of computing efficiency that generalizes across model scales and deployment contexts. But Max-ACC optimization is the best when high ACC is demanded.

\subsection{One Null Case for Knee Detection}
In one configuration in case 1, we failed to identify a knee point because the feasible Pareto set under the imposed constraints collapsed to at most two non-dominated points, rendering curvature-based knee detection ill-defined. In such cases, utopia-closet or cube-volume can serve an alternative approach to balance efficiency.

\section{Discussion}
\subsection{Toward 3D Optimization from 1D and 2D}
Conventional approaches to inference scaling typically relied on 1D heuristics, such as executing a fixed number of reasoning passes (e.g., 16 passes), or on 2D tradeoffs between performance and compute scale. While these methods improved accuracy, they failed to capture the constraints of latency and cost that are essential for AI inference. Because real-world inference should operate within given cost and latency budgets, ignoring these constraints leads to impractical inference. The proposed 3D MOO overcomes the limitation of past approaches by jointly optimizing accuracy, cost, and latency. This method reveals new feasible regions that 1D and 2D optimization fail to capture. The 3D objective space provides an opportunity for a deployment-aware inference optimization.

\subsection{Efficiency through Scaling with Parallelism}
The results demonstrate that additional compute allocation can compensate the performance for the smaller size. When parallelism is available, smaller models can achieve comparable or even superior accuracy to larger architectures with a lower cost and latency. For example, GPT-5~Nano with 56 inference paths showed ACC of 0.99 (\$0.05, 30.1s), which is greater than GPT-5 with 24 inference paths that achieved ACC of 0.98 (\$0.52, 64.5s) under the knee-point selection criterion in Scenario 3. This result suggests that inference scaling offers an efficient path for small models, especially when parallel inference is available.

\subsection{Knee-Point Optimization for the Best Balance}
Among MOOs, knee-point selection consistently achieved the most balanced efficiency in the feasible space. The Knee MOO identifies the point of maximum curvature on the Pareto frontier, showing the relative efficiency in tri-objectives. While it could be fail to be identified for the linearity of the frontiers that does not show the knee point, it can serve as the most practical solution except for the case that needs the maximal accuracy. Regarding the inference for the minimal cost and the best latency, the solution is always inference with a single path.

\subsection{Limitations and Future Work}
This study relies on stochastic simulation rather than LLM deployments. Thus, it abstracts away system-level factors such as GPU utilization, memory contention, and network queuing delays. Real-world AI infrastructures often exhibit non-Gaussian variability due to heterogeneous hardware, fluctuating energy conditions, and network overhead, all of which complicate inference MOO for further dimensions and higher dimensions. Incorporating empirical traces from AI infrastructure environments would further validate the proposed framework.

Also, the proposed AI inference MOO framework is critical not only for single-agent models but also for multi-agent systems, where multiple agents and tools compound the cost and latency overhead, further amplifying the need for efficient inference scaling. Moreover, as dimensions such as energy consumption, safety constraints, and thermal limits become increasingly relevant, the framework can be extended beyond 3D to support higher-dimensional MOO. Such a generalized formulation would enable jointly optimizing energy, safety, and heat dissipation on top of the proposed 3D space will support deployment-aware reasoning for real-world AI systems.

Future work could extend our framework to infrastructure-aware optimization. This includes integrating hardware metrics from GPUs and Kubernetes-based orchestration to align inference scaling with real-time resource allocation. The optimization layer could operate as an adaptive router that dynamically selects the model and inference scale based on complexity and system load. Such integration would enable real-world MOO across heterogeneous pools, improving inference efficiency on a large scale.

\section{Conclusion}
We reframe AI inference scaling as a multidimensional optimization problem that considers accuracy, cost, and latency in one optimization to overcome the limitations of 1D and 2D optimization. By integrating MC simulations with MOO, the proposed framework enables constraint-aware inference scaling. Across diverse model families and scenarios, the result revealed that knee-optimality from the Pareto frontiers makes the scale that shows a relative efficiency among others. Also, we found that smaller models could show a better performance than larger models with a lower cost and latency. Beyond these findings, this framework establishes a foundation for advanced inference systems that respond to operational conditions. Embedding MOO into orchestration layers will enable efficient inference across heterogeneous AI models and diverse environments. As foundation models and agentic systems continue to expand, principled inference scaling will become essential to deploy AI efficiently.


\section{Acknowledgments}
This project is a part of Cloudera's Accelerators for ML Projects (AMPs). The AMP is accessible via \url{https://github.com/cloudera/CAI_AMP_Inference_Scaling_Optimization}.
Thanks to  Nashua Springberry and Michael Schuler for constructive comments on the design and programming for the simulation.

\bibliography{aaai2026bib.bib}

@article{snell2024scaling,
  title={Scaling llm test-time compute optimally can be more effective than scaling model parameters},
  author={Snell, Charlie and Lee, Jaehoon and Xu, Kelvin and Kumar, Aviral},
  journal={arXiv preprint arXiv:2408.03314},
  year={2024}
}

@article{wu2024inference,
  title={Inference scaling laws: An empirical analysis of compute-optimal inference for problem-solving with language models},
  author={Wu, Yangzhen and Sun, Zhiqing and Li, Shanda and Welleck, Sean and Yang, Yiming},
  journal={arXiv preprint arXiv:2408.00724},
  year={2024}
}

@article{park2025know,
  title={Know What You Don't Know: Uncertainty Calibration of Process Reward Models},
  author={Park, Young-Jin and Greenewald, Kristjan and Alim, Kaveh and Wang, Hao and Azizan, Navid},
  journal={arXiv preprint arXiv:2506.09338},
  year={2025}
}

@article{bian2025scaling,
  title={Scaling Inference-Efficient Language Models},
  author={Bian, Song and Yan, Minghao and Venkataraman, Shivaram},
  journal={arXiv preprint arXiv:2501.18107},
  year={2025}
}

@article{wang2022self,
  title={Self-consistency improves chain of thought reasoning in language models},
  author={Wang, Xuezhi and Wei, Jason and Schuurmans, Dale and Le, Quoc and Chi, Ed and Narang, Sharan and Chowdhery, Aakanksha and Zhou, Denny},
  journal={arXiv preprint arXiv:2203.11171},
  year={2022}
}

@article{jin2025energy,
  title={The Energy Cost of Reasoning: Analyzing Energy Usage in LLMs with Test-time Compute},
  author={Jin, Yunho and Wei, Gu-Yeon and Brooks, David},
  journal={arXiv preprint arXiv:2505.14733},
  year={2025}
}

@article{sadhukhan2025kinetics,
  title={Kinetics: Rethinking Test-Time Scaling Laws},
  author={Sadhukhan, Ranajoy and Chen, Zhuoming and Zheng, Haizhong and Zhou, Yang and Strubell, Emma and Chen, Beidi},
  journal={arXiv preprint arXiv:2506.05333},
  year={2025}
}

@article{nvidia2025balancing,
  title={The Art of Balancing AI Inference Cost and  Performance},
  author={NVIDIA},
  journal={https://nvdam.widen.net/s/hrprjhtmm9/the-it-leaders-guide-to-ai-inference-and-performance},
  year={2025}
}

@article{aubrey2025maximizeai,
  title        = {How the Economics of Inference Can Maximize AI Value},
  author       = {Aubrey, Kyle},
  year         = {2025},
  journal      = {NVIDIA Technical Blog},
  note         = {Available at \url{https://blogs.nvidia.com/blog/ai-inference-economics/}},
}

@article{harris2025thinksmart,
  title        = {Think SMART: How to Optimize AI Factory Inference Performance},
  author       = {Harris, Dion},
  year         = {2025},
  journal      = {NVIDIA Technical Blog},
  note         = {Available at \url{https://blogs.nvidia.com/blog/think-smart-optimize-ai-factory-inference-performance/}},
}

@article{brown2024large,
  title={Large language monkeys: Scaling inference compute with repeated sampling},
  author={Brown, Bradley and Juravsky, Jordan and Ehrlich, Ryan and Clark, Ronald and Le, Quoc V and R{\'e}, Christopher and Mirhoseini, Azalia},
  journal={arXiv preprint arXiv:2407.21787},
  year={2024}
}

\onecolumn
\appendix
\section{Model Configuration and Hyperparameters}\label{appendix:model_config}
To evaluate the proposed inference scaling framework, we define a set of representative model configurations that approximate contemporary large language models (LLMs) across different families and scales. These configurations are not tied to proprietary systems but
serve as controllable abstractions for stochastic simulation. Each model specifies per-token costs, latencies, and accuracy distributions used in the Monte Carlo analysis. Costs are expressed in U.S.\ dollars per million tokens, and latency values represent seconds per input ($t_\mathrm{in}$) or output ($t_\mathrm{out}$) token. Unless otherwise stated, the
parallelism factor is fixed at $P=4$. Table~\ref{tab:model_configs} summarizes all hyperparameters.

\begin{table}[H]
\centering

\begin{tabular}{lcccccccc}
\toprule
\textbf{Model} &
$c_\mathrm{in}$ &
$c_\mathrm{out}$ &
$t_\mathrm{in}$ &
$t_\mathrm{out}$ &
$\mu_{L_\mathrm{in}}$ &
$\mu_{L_\mathrm{out}}$ &
$\text{Accuracy Mean} \pm \text{Std}$ &
$P_\text{default}$ \\
\midrule
GPT-5 & 1.25e-6 & 10.00e-6 & 0.0005 & 0.0050 & 1024 & 2048 & 0.94 $\pm$ 0.02 & 4 \\
GPT-5 Mini & 0.25e-6 & 2.00e-6 & 0.00025 & 0.0020 & 1024 & 2048 & 0.92 $\pm$ 0.03 & 4 \\
GPT-5 Nano & 0.05e-6 & 0.40e-6 & 0.00010 & 0.0010 & 1024 & 2048 & 0.91 $\pm$ 0.04 & 4 \\
Nvidia Nemotron Ultra 253B & 0.90e-6 & 2.80e-6 & 0.0010 & 0.0100 & 1024 & 2048 & 0.93 $\pm$ 0.05 & 4 \\
Nvidia Nemotron H 47B & 0.40e-6 & 1.50e-6 & 0.0004 & 0.0040 & 1024 & 2048 & 0.92 $\pm$ 0.06 & 4 \\
Nvidia Nemotron Nano 9B v2 & 0.20e-6 & 1.00e-6 & 0.00012 & 0.0012 & 1024 & 2048 & 0.91 $\pm$ 0.07 & 4 \\
Qwen3-Max & 0.90e-6 & 2.40e-6 & 0.0008 & 0.0080 & 1024 & 2048 & 0.90 $\pm$ 0.04 & 4 \\
Qwen3-Next-80B-A3B & 0.50e-6 & 1.25e-6 & 0.0004 & 0.0040 & 1024 & 2048 & 0.89 $\pm$ 0.05 & 4 \\
Qwen3-30B-A3B & 0.35e-6 & 0.90e-6 & 0.00025 & 0.0020 & 1024 & 2048 & 0.88 $\pm$ 0.06 & 4 \\
\bottomrule
\end{tabular}
\caption{Representative model configurations used for stochastic simulation.
Costs are given in USD per million tokens.
Latency values denote seconds per token for input ($t_\mathrm{in}$) and output ($t_\mathrm{out}$).
All models assume a default parallelism of $P=4$.}
\label{tab:model_configs}
\end{table}

These parameters ensure that the optimal inference scale $k^\star$ under the default
budget constraints ($C_{\max}=0.50$, $T_{\max}=60$ s) typically falls within a practical
range ($k \approx 10$–$20$), allowing interpretable comparisons across model families
while maintaining computational realism.

\section{Simulations}\label{full_simulations}

\subsection{Scenario 1: Low Cost and Time Budget (Essay Feedback)}\label{appendix:scenario_1}
This scenario represents lightweight applications such as automated essay evaluation or short-form feedback generation,
where users prioritize quick turnaround and low cost while maintaining reasonable accuracy.
We set the budget constraints to a maximum cost of \$0.50 and a latency limit of 60 seconds,
with a minimum acceptable accuracy of $A_{\min}=0.93$.
Under these deployment limits, four optimization strategies—accuracy-maximization, cube-optimal
(multiplicative balance), utopia-closest, and knee-point—are evaluated across nine representative models.

Table~\ref{tab:scenario1} summarizes the simulation results.
The knee-point method consistently achieves the best efficiency across most models,
delivering near-maximal accuracy with notably reduced cost and latency.
Cube- and utopia-based selections favor minimal resource use but sacrifice accuracy,
while accuracy-maximization frequently over-computes within the same budget.
Overall, knee-point optimization provides the most balanced solution for latency-sensitive
or cost-limited settings such as real-time educational feedback.

\begin{table}[H]
\centering
\begin{tabular}{lcccc}
\toprule
\textbf{Model} & \textbf{Accuracy-Optimal} & \textbf{Cube-Optimal} & \textbf{Utopia-Closest} & \textbf{Knee-Point} \\
\midrule
G-5 & k=16 / 0.98 / 0.35 / 43.1 & k=4 / 0.96 / 0.09 / 10.8 & k=4 / 0.96 / 0.09 / 10.8 & k=8 / 0.97 / 0.17 / 21.5 \\
G-5M & k=52 / 0.99 / 0.23 / 56.6 & k=4 / 0.95 / 0.02 / 4.4 & k=4 / 0.95 / 0.02 / 4.4 & k=40 / 0.98 / 0.17 / 43.5 \\
G-5N & k=108 / 1.00 / 0.09 / 58.1 & k=4 / 0.95 / 0.00 / 2.2 & k=4 / 0.95 / 0.00 / 2.2 & k=56 / 0.99 / 0.05 / 30.1 \\
N-253B & k=8 / 0.99 / 0.05 / 42.9 & k=4 / 0.97 / 0.03 / 21.5 & k=4 / 0.97 / 0.03 / 21.5 & – \\
N-47B & k=24 / 1.00 / 0.08 / 51.6 & k=4 / 0.97 / 0.01 / 8.6 & k=4 / 0.97 / 0.01 / 8.6 & k=8 / 0.99 / 0.03 / 17.2 \\
N-9B & k=52 / 1.00 / 0.12 / 33.5 & k=4 / 0.97 / 0.01 / 2.6 & k=4 / 0.97 / 0.01 / 2.6 & k=48 / 1.00 / 0.11 / 31.0 \\
Q3-M & k=12 / 0.96 / 0.07 / 51.6 & k=4 / 0.94 / 0.02 / 17.2 & k=4 / 0.94 / 0.02 / 17.2 & k=8 / 0.96 / 0.05 / 34.4 \\
Q3-80B & k=24 / 0.98 / 0.07 / 51.6 & k=4 / 0.94 / 0.01 / 8.6 & k=4 / 0.94 / 0.01 / 8.6 & k=8 / 0.96 / 0.03 / 17.2 \\
Q3-30B & k=52 / 1.00 / 0.11 / 56.6 & k=4 / 0.94 / 0.01 / 4.4 & k=4 / 0.94 / 0.01 / 4.4 & k=48 / 0.99 / 0.11 / 52.2 \\
\bottomrule
\end{tabular}
\caption{
Scenario~1 (Essay Feedback). 
Optimal configurations across nine models under $C_{\max}=0.50$, $T_{\max}=60$~s, and $A_{\min}=0.93$.
Each cell shows $k$ / accuracy / total cost (\$) / total latency (s). 
Model abbreviations: G--5 (GPT--5), G--5M (GPT--5 Mini), G--5N (GPT--5 Nano), 
N--253B (Nemotron Ultra 253B), N--47B (Nemotron H 47B), N--9B (Nemotron Nano 9B v2),
Q3--M (Qwen3-Max), Q3--80B (Qwen3-Next-80B-A3B), and Q3--30B (Qwen3-30B-A3B). 
Standard deviations (omitted for brevity) were all within approximately $\pm 0.01$.}
\label{tab:scenario1}
\end{table}

\subsection{Scenario 2: High Cost and Time Budget (Medical AI)}
This scenario reflects mission-critical domains such as medical diagnostics or clinical report generation,
where users prioritize maximum accuracy and reliability over latency or cost considerations.
Here, the system operates under relaxed constraints with a maximum cost of \$0.95 and a latency budget of 3600 seconds,
while requiring a minimum target accuracy of $A_{\min}=0.98$.
Under these generous budgets, we again evaluate four optimization strategies—accuracy-maximization, cube-optimal,
utopia-closest, and knee-point—across nine representative model configurations.

Table~\ref{tab:scenario2} summarizes the results.
Accuracy-maximization consistently delivers the highest absolute accuracy, making it preferable in precision-critical settings such as medical AI.
However, the knee-point approach achieves nearly identical accuracy at substantially lower computational cost and latency,
demonstrating superior efficiency when resource constraints or throughput considerations are present.
Cube- and utopia-based methods remain the most cost-efficient but trade off small amounts of accuracy.
Overall, these findings suggest that while accuracy-maximization remains ideal for clinical-grade precision,
knee-point optimization provides a more balanced and deployable alternative.

\begin{table}[H]
\centering
\begin{tabular}{lcccc}
\toprule
\textbf{Model} & \textbf{Accuracy-Optimal} & \textbf{Cube-Optimal} & \textbf{Utopia-Closest} & \textbf{Knee-Point} \\
\midrule
G--5 & k=36 / 0.98 / 0.78 / 96.7 & k=28 / 0.98 / 0.61 / 75.2 & k=28 / 0.98 / 0.61 / 75.2 & k=32 / 0.98 / 0.70 / 86.0 \\
G--5M & k=128 / 0.99 / 0.56 / 139.3 & k=32 / 0.98 / 0.14 / 34.8 & k=32 / 0.98 / 0.14 / 34.8 & k=64 / 0.99 / 0.28 / 69.7 \\
G--5N & k=128 / 1.00 / 0.11 / 68.8 & k=16 / 0.98 / 0.01 / 8.6 & k=16 / 0.98 / 0.01 / 8.6 & k=56 / 0.99 / 0.05 / 30.1 \\
N--253B & k=72 / 1.00 / 0.48 / 386.9 & k=8 / 0.99 / 0.05 / 42.9 & k=8 / 0.99 / 0.05 / 42.9 & k=48 / 1.00 / 0.32 / 258.0 \\
N--47B & k=52 / 1.00 / 0.18 / 111.8 & k=8 / 0.99 / 0.03 / 17.2 & k=8 / 0.99 / 0.03 / 17.2 & k=48 / 1.00 / 0.17 / 103.2 \\
N--9B & k=52 / 1.00 / 0.12 / 33.5 & k=8 / 0.99 / 0.02 / 5.2 & k=8 / 0.99 / 0.02 / 5.2 & k=48 / 1.00 / 0.11 / 31.0 \\
Q3--M & k=128 / 1.00 / 0.75 / 550.5 & k=32 / 0.98 / 0.19 / 137.6 & k=32 / 0.98 / 0.19 / 137.6 & k=56 / 0.99 / 0.33 / 240.8 \\
Q3--80B & k=128 / 1.00 / 0.39 / 275.3 & k=24 / 0.98 / 0.07 / 51.6 & k=24 / 0.98 / 0.07 / 51.6 & k=68 / 1.00 / 0.21 / 146.3 \\
Q3--30B & k=128 / 1.00 / 0.28 / 139.3 & k=16 / 0.98 / 0.04 / 17.4 & k=16 / 0.98 / 0.04 / 17.4 & k=68 / 1.00 / 0.15 / 74.0 \\
\bottomrule
\end{tabular}
\caption{
Scenario~2 (Medical AI). 
Optimal configurations across nine models under $C_{\max}=0.95$, $T_{\max}=3600$~s, and $A_{\min}=0.98$.
Each cell shows $k$ / accuracy / total cost (\$) / total latency (s). 
Model abbreviations follow Table~\ref{tab:scenario1}. 
Standard deviations (omitted for brevity) were all within approximately $\pm 0.01$.}
\label{tab:scenario2}
\end{table}

\subsection{Scenario 3: Moderate Cost and Strict Latency (Proposal Writing)}
This scenario models creative or document-generation tasks such as research proposal drafting or long-form text refinement,
where users can tolerate longer inference durations but still operate under limited cost budgets.
We set a moderate cost constraint of \$0.65 and a latency limit of 1800 seconds,
with a minimum acceptable accuracy of $A_{\min}=0.96$.
Under these deployment conditions, four optimization strategies—accuracy-maximization, cube-optimal,
utopia-closest, and knee-point—are compared across nine representative language models.

Table~\ref{tab:scenario3} presents the results.
The knee-point method again demonstrates superior efficiency, consistently achieving accuracy close to the
accuracy-optimal solution while reducing total cost and latency by up to 40–70\%.
Accuracy-maximization delivers the best possible accuracy but often over-computes under the same cost limit,
while cube- and utopia-based methods minimize resource use but underperform in accuracy-sensitive settings.
Overall, knee-point optimization remains the most balanced approach for moderate-cost,
latency-constrained tasks such as real-time writing or interactive drafting systems.

\begin{table}[H]
\centering
\begin{tabular}{lcccc}
\toprule
\textbf{Model} & \textbf{Accuracy-Optimal} & \textbf{Cube-Optimal} & \textbf{Utopia-Closest} & \textbf{Knee-Point} \\
\midrule
G--5 & k=28 / 0.98 / 0.61 / 75.2 & k=8 / 0.97 / 0.17 / 21.5 & k=8 / 0.97 / 0.17 / 21.5 & k=24 / 0.98 / 0.52 / 64.5 \\
G--5M & k=128 / 0.99 / 0.56 / 139.3 & k=8 / 0.96 / 0.04 / 8.7 & k=8 / 0.96 / 0.04 / 8.7 & k=68 / 0.99 / 0.30 / 74.0 \\
G--5N & k=128 / 1.00 / 0.11 / 68.8 & k=16 / 0.98 / 0.01 / 8.6 & k=16 / 0.98 / 0.01 / 8.6 & k=56 / 0.99 / 0.05 / 30.1 \\
N--253B & k=72 / 1.00 / 0.48 / 386.9 & k=4 / 0.97 / 0.03 / 21.5 & k=4 / 0.97 / 0.03 / 21.5 & k=48 / 1.00 / 0.32 / 258.0 \\
N--47B & k=52 / 1.00 / 0.18 / 111.8 & k=4 / 0.97 / 0.01 / 8.6 & k=4 / 0.97 / 0.01 / 8.6 & k=48 / 1.00 / 0.17 / 103.2 \\
N--9B & k=52 / 1.00 / 0.12 / 33.5 & k=8 / 0.99 / 0.02 / 5.2 & k=8 / 0.99 / 0.02 / 5.2 & k=48 / 1.00 / 0.11 / 31.0 \\
Q3--M & k=108 / 1.00 / 0.63 / 464.6 & k=12 / 0.96 / 0.07 / 51.6 & k=12 / 0.96 / 0.07 / 51.6 & k=56 / 0.99 / 0.33 / 240.8 \\
Q3--80B & k=128 / 1.00 / 0.39 / 275.3 & k=12 / 0.97 / 0.04 / 25.8 & k=12 / 0.97 / 0.04 / 25.8 & k=68 / 1.00 / 0.21 / 146.3 \\
Q3--30B & k=128 / 1.00 / 0.28 / 139.3 & k=8 / 0.96 / 0.02 / 8.7 & k=8 / 0.96 / 0.02 / 8.7 & k=68 / 1.00 / 0.15 / 74.0 \\
\bottomrule
\end{tabular}
\caption{
Scenario~3 (Proposal Writing). 
Optimal configurations across nine models under $C_{\max}=0.65$, $T_{\max}=1800$~s, and $A_{\min}=0.96$.
Each cell shows $k$ / accuracy / total cost (\$) / total latency (s). 
Model abbreviations follow Table~\ref{tab:scenario1}. 
Standard deviations (omitted for brevity) were all within approximately $\pm 0.01$.}
\label{tab:scenario3}
\end{table}

\section{Optimization Results for Inference Scaling across Models and Scenarios}\label{appendix:visualizations}

This section provides 3D and 2D visualizations of the optimization process of the simulation. It includes the simulation for three scenarios across three models with large, medium, and small model size. The result for the sample GPT-5 simulation has been noted in the main body of the paper. 

\subsection{GPT5-mini}
This shows the MOO from the GPT-mini case across three scenarios.

\begin{figure}[H]
  \centering
  \begin{subfigure}{0.32\textwidth}
    \includegraphics[width=\linewidth]{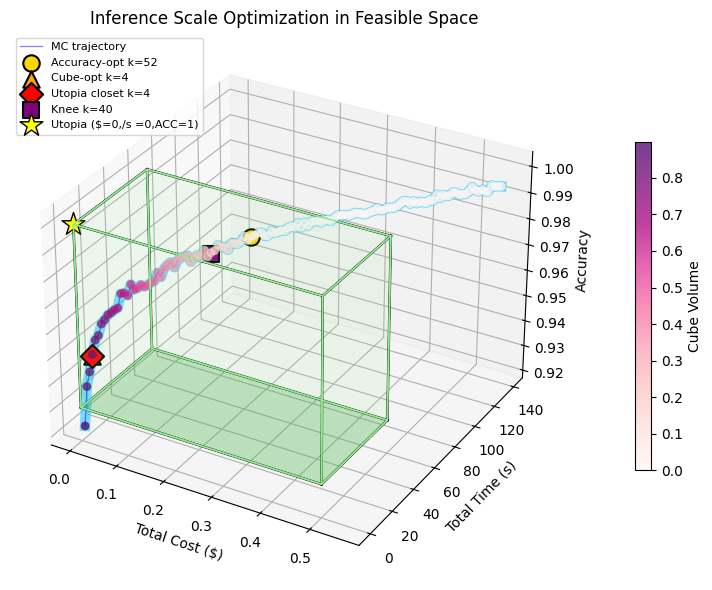}
    \caption{3D MOO results for Case~1.}
  \end{subfigure}
  \begin{subfigure}{0.32\textwidth}
    \includegraphics[width=\linewidth]{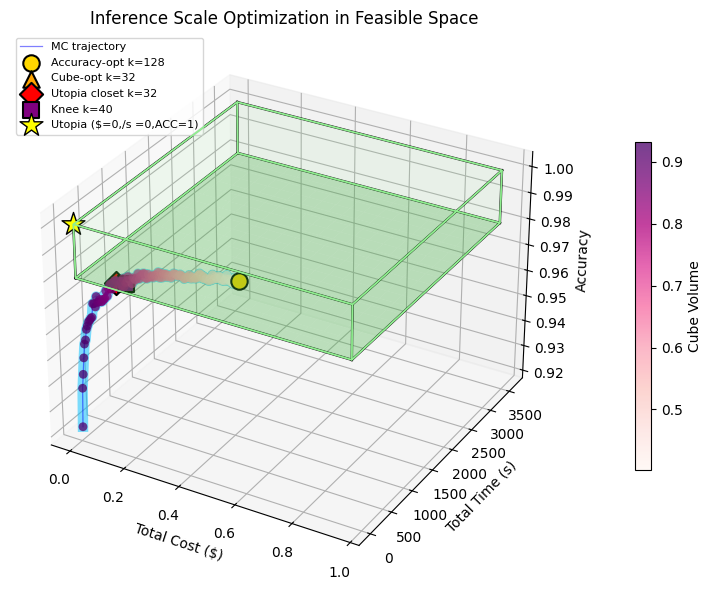}
    \caption{3D MOO results for Case~2.}
  \end{subfigure}
  \begin{subfigure}{0.32\textwidth}
    \includegraphics[width=\linewidth]{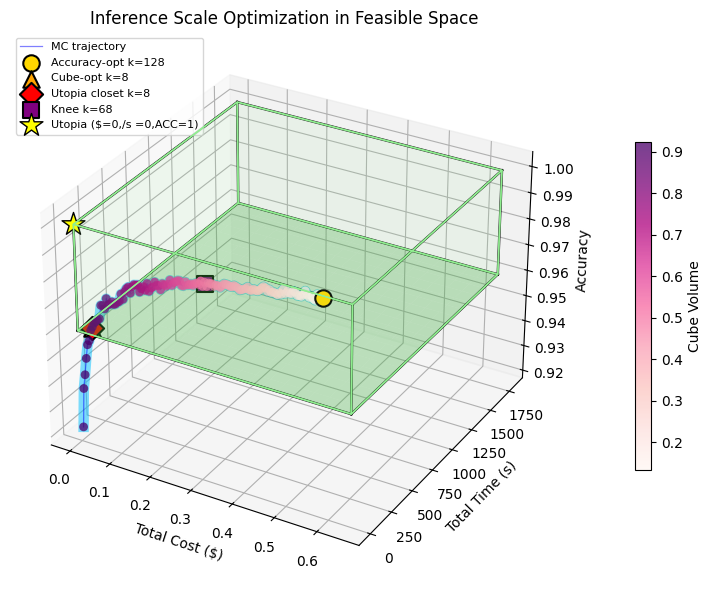}
    \caption{3D MOO results for Case~3.}
  \end{subfigure}
    \caption{
    Inference-scaling optimization results for GPT5-mini across three simulated scenarios. 
    (a--c) 3D feasible cubes in cost--time--accuracy space with constraint planes at \(C_{\max}\), \(T_{\max}\), and \(A_{\min}\). 
    Markers denote optimization criteria (\(\bullet\) Accuracy-Optimal, \(\blacktriangle\) Cube-Optimal, \(\blacklozenge\) Utopia-Closest, \(\blacksquare\) Knee-Point); colors indicate cube volume (larger is better), and the yellow star marks the utopia point.}
\end{figure}

\begin{figure}[H]
  \centering
  \begin{subfigure}{0.32\textwidth}
    \includegraphics[width=\linewidth]{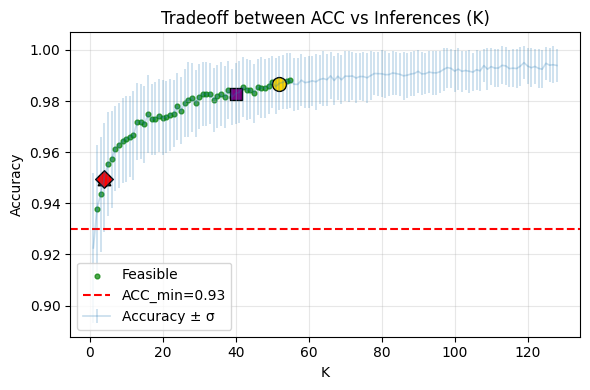}
    \caption{2D ACC--K with optimality for Case~1.}
  \end{subfigure}
  \begin{subfigure}{0.32\textwidth}
    \includegraphics[width=\linewidth]{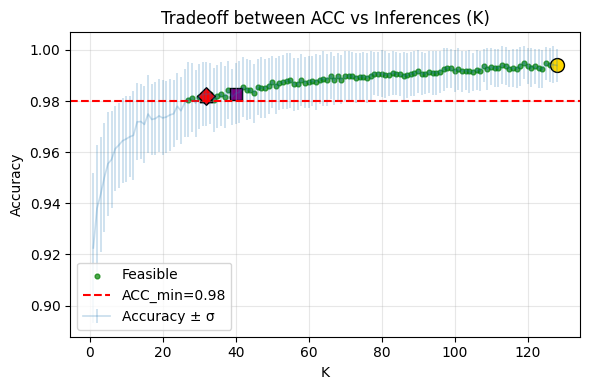}
    \caption{2D ACC--K with optimality for Case~2.}
  \end{subfigure}
  \begin{subfigure}{0.32\textwidth}
    \includegraphics[width=\linewidth]{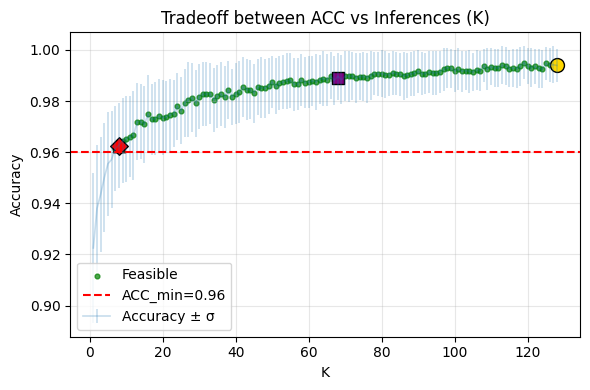}
    \caption{2D ACC--K with optimality for Case~3.}
  \end{subfigure}
    \caption{Accuracy-inference compute (K) trade-offs across three scenarios. These show Monte Carlo means with CI = 95\%; the red dashed line indicates \(A_{\min}\). Distinct operational priorities yield different optimal inference scales \(k^\star\).}
\end{figure}

\subsection{GPT5-nano}

This shows the MOO from the GPT-nano case across three scenarios.

\begin{figure}[H]
  \centering
  \begin{subfigure}{0.32\textwidth}
    \includegraphics[width=\linewidth]{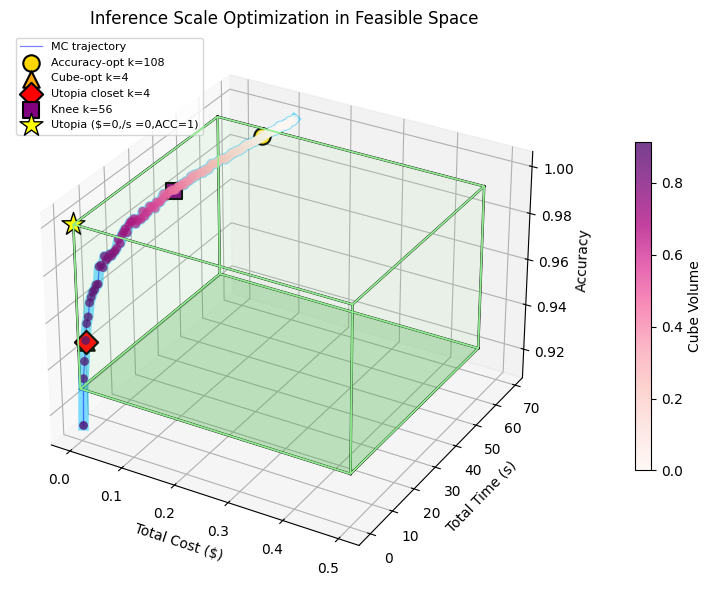}
    \caption{3D MOO results for Case~1.}
  \end{subfigure}
  \begin{subfigure}{0.32\textwidth}
    \includegraphics[width=\linewidth]{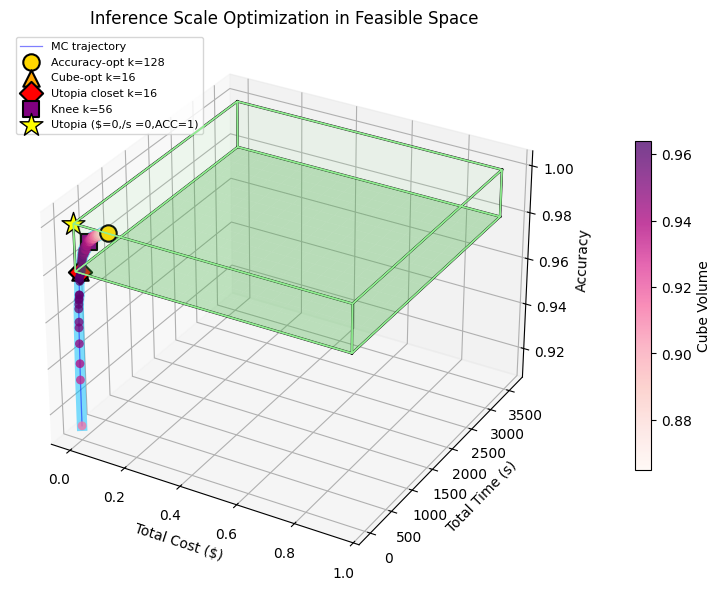}
    \caption{3D MOO results for Case~2.}
  \end{subfigure}
  \begin{subfigure}{0.32\textwidth}
    \includegraphics[width=\linewidth]{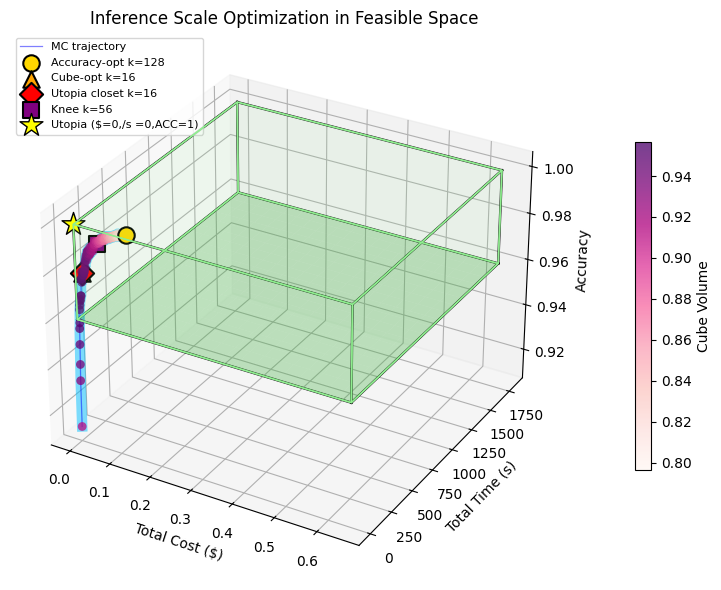}
    \caption{3D MOO results for Case~3.}
  \end{subfigure}
    \caption{
    Inference-scaling optimization results for GPT5-nano across three simulated scenarios. 
    (a--c) 3D feasible cubes in cost--time--accuracy space with constraint planes at \(C_{\max}\), \(T_{\max}\), and \(A_{\min}\). 
    Markers denote optimization criteria (\(\bullet\) Accuracy-Optimal, \(\blacktriangle\) Cube-Optimal, \(\blacklozenge\) Utopia-Closest, \(\blacksquare\) Knee-Point); colors indicate cube volume (larger is better), and the yellow star marks the utopia point.}
\end{figure}

\begin{figure}[H]
  \centering
  \begin{subfigure}{0.32\textwidth}
    \includegraphics[width=\linewidth]{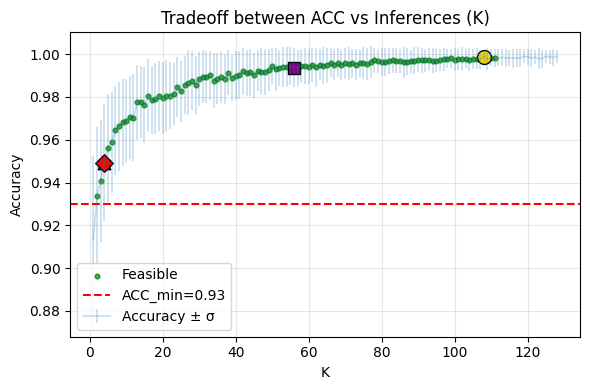}
    \caption{2D ACC--K with optimality for Case~1.}
  \end{subfigure}
  \begin{subfigure}{0.32\textwidth}
    \includegraphics[width=\linewidth]{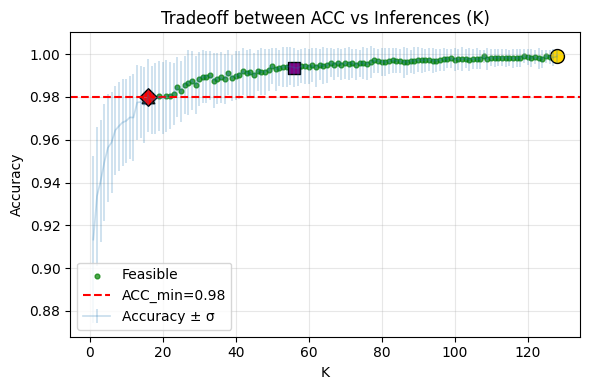}
    \caption{2D ACC--K with optimality for Case~2.}
  \end{subfigure}
  \begin{subfigure}{0.32\textwidth}
    \includegraphics[width=\linewidth]{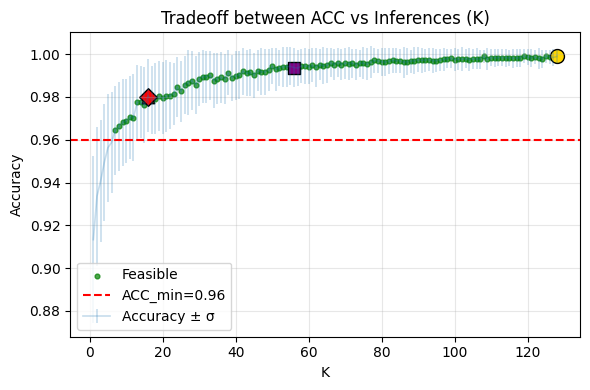}
    \caption{2D ACC--K with optimality for Case~3.}
  \end{subfigure}
    \caption{Accuracy-inference compute (K) trade-offs across three scenarios. These show Monte Carlo means with CI = 95\%; the red dashed line indicates \(A_{\min}\). Distinct operational priorities yield different optimal inference scales \(k^\star\).}
\end{figure}

\subsection{Nvidia Nemotron Ultra 253B}
This shows the MOO from the Nvidia Nemotron Ultra 253B case across three scenarios.

\begin{figure}[H]
  \centering
  \begin{subfigure}{0.32\textwidth}
    \includegraphics[width=\linewidth]{gpt5-mini_case1_3d.png}
    \caption{3D MOO results for Case~1.}
  \end{subfigure}
  \begin{subfigure}{0.32\textwidth}
    \includegraphics[width=\linewidth]{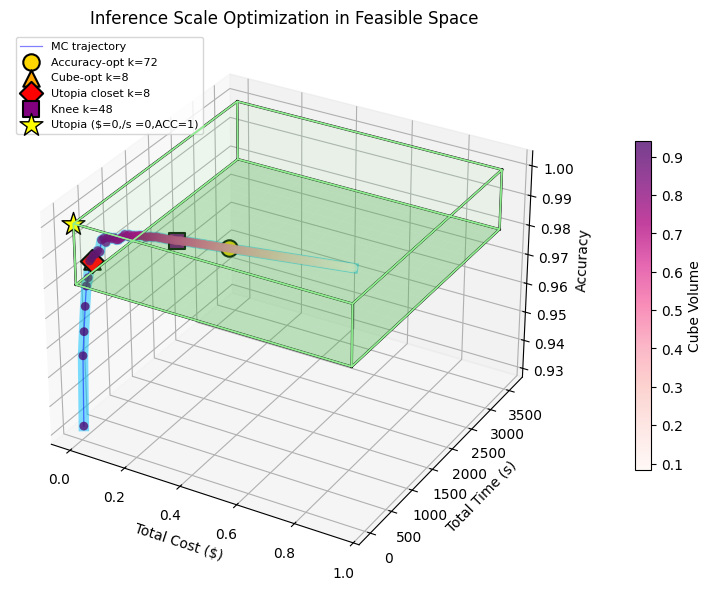}
    \caption{3D MOO results for Case~2.}
  \end{subfigure}
  \begin{subfigure}{0.32\textwidth}
    \includegraphics[width=\linewidth]{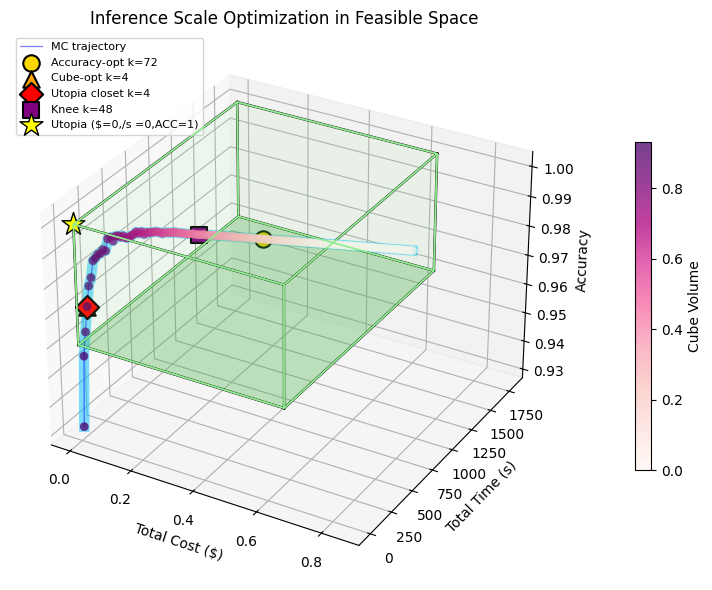}
    \caption{3D MOO results for Case~3.}
  \end{subfigure}
    \caption{
    Inference-scaling optimization results for Nvidia Nemotron Ultra 253B across three simulated scenarios. 
    (a--c) 3D feasible cubes in cost--time--accuracy space with constraint planes at \(C_{\max}\), \(T_{\max}\), and \(A_{\min}\). 
    Markers denote optimization criteria (\(\bullet\) Accuracy-Optimal, \(\blacktriangle\) Cube-Optimal, \(\blacklozenge\) Utopia-Closest, \(\blacksquare\) Knee-Point); colors indicate cube volume (larger is better), and the yellow star marks the utopia point.}
\end{figure}

\begin{figure}[H]
  \centering
  \begin{subfigure}{0.32\textwidth}
    \includegraphics[width=\linewidth]{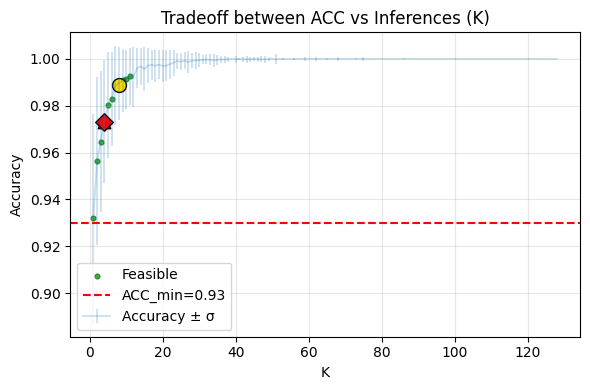}
    \caption{2D ACC--K with optimality for Case~1.}
  \end{subfigure}
  \begin{subfigure}{0.32\textwidth}
    \includegraphics[width=\linewidth]{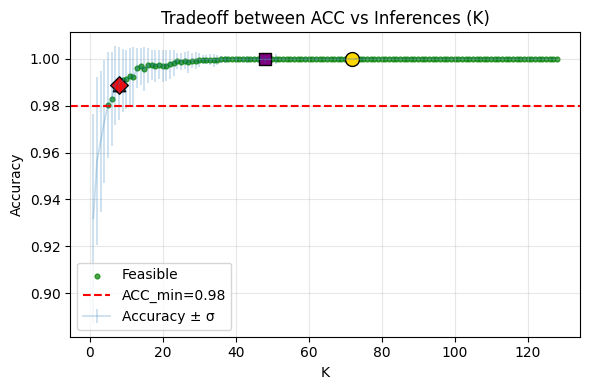}
    \caption{2D ACC--K with optimality for Case~2.}
  \end{subfigure}
  \begin{subfigure}{0.32\textwidth}
    \includegraphics[width=\linewidth]{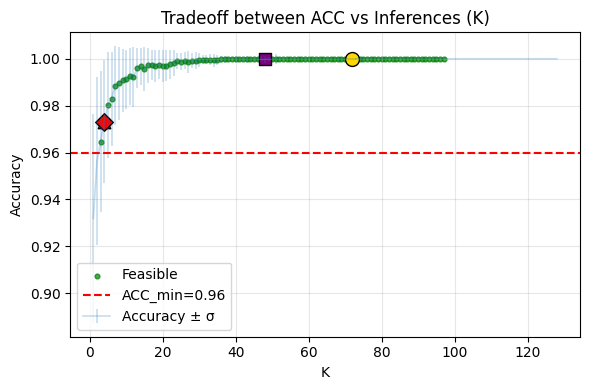}
    \caption{2D ACC--K with optimality for Case~3.}
  \end{subfigure}
    \caption{Accuracy-inference compute (K) trade-offs across three scenarios. These show Monte Carlo means with CI = 95\%; the red dashed line indicates \(A_{\min}\). Distinct operational priorities yield different optimal inference scales \(k^\star\).}
\end{figure}

\subsection{Nvidia Nemotron H 47B}
This shows the MOO from the Nvidia Nemotron H 47B case across three scenarios.

\begin{figure}[H]
  \centering
  \begin{subfigure}{0.32\textwidth}
    \includegraphics[width=\linewidth]{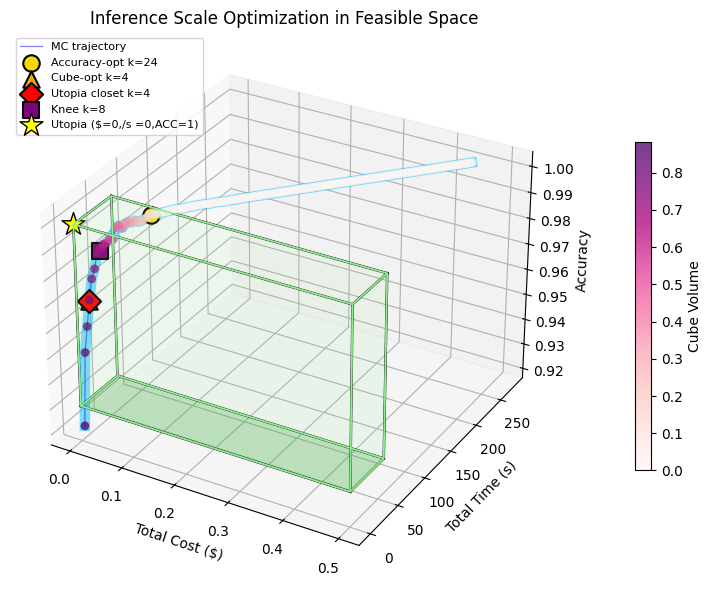}
    \caption{3D MOO results for Case~1.}
  \end{subfigure}
  \begin{subfigure}{0.32\textwidth}
    \includegraphics[width=\linewidth]{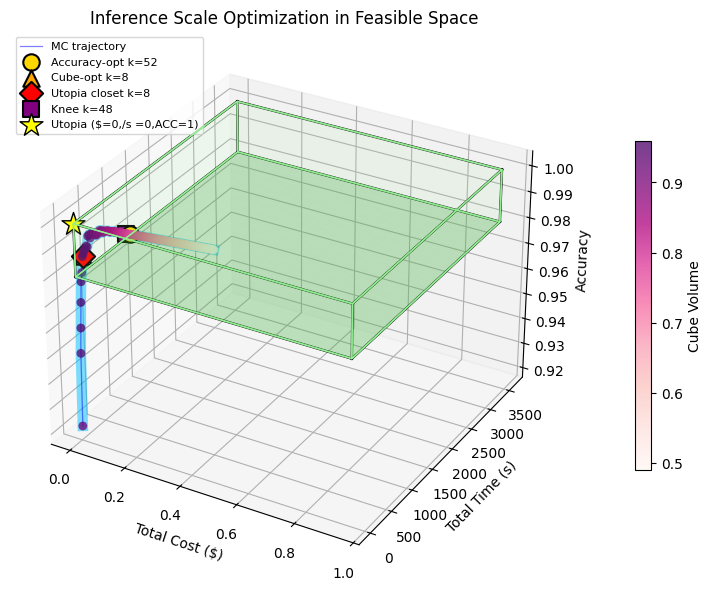}
    \caption{3D MOO results for Case~2.}
  \end{subfigure}
  \begin{subfigure}{0.32\textwidth}
    \includegraphics[width=\linewidth]{nemo_med_case2_3d.png}
    \caption{3D MOO results for Case~3.}
  \end{subfigure}
    \caption{
    Inference-scaling optimization results for Nvidia Nemotron H 47B across three simulated scenarios. 
    (a--c) 3D feasible cubes in cost--time--accuracy space with constraint planes at \(C_{\max}\), \(T_{\max}\), and \(A_{\min}\). 
    Markers denote optimization criteria (\(\bullet\) Accuracy-Optimal, \(\blacktriangle\) Cube-Optimal, \(\blacklozenge\) Utopia-Closest, \(\blacksquare\) Knee-Point); colors indicate cube volume (larger is better), and the yellow star marks the utopia point.}
\end{figure}

\begin{figure}[H]
  \centering
  \begin{subfigure}{0.32\textwidth}
    \includegraphics[width=\linewidth]{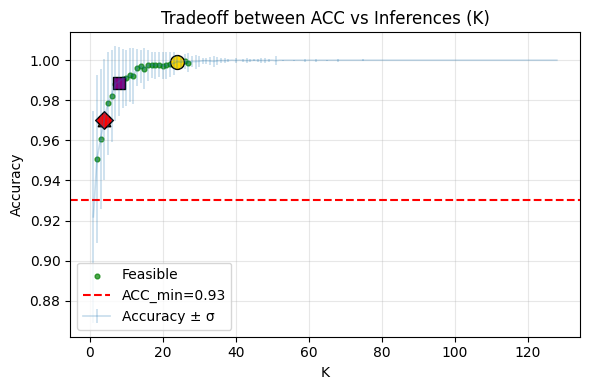}
    \caption{2D ACC--K with optimality for Case~1.}
  \end{subfigure}
  \begin{subfigure}{0.32\textwidth}
    \includegraphics[width=\linewidth]{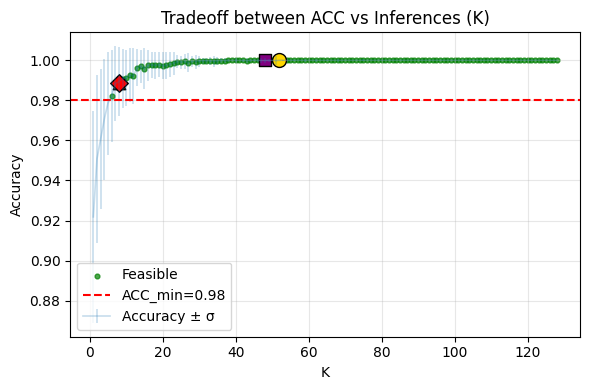}
    \caption{2D ACC--K with optimality for Case~2.}
  \end{subfigure}
  \begin{subfigure}{0.32\textwidth}
    \includegraphics[width=\linewidth]{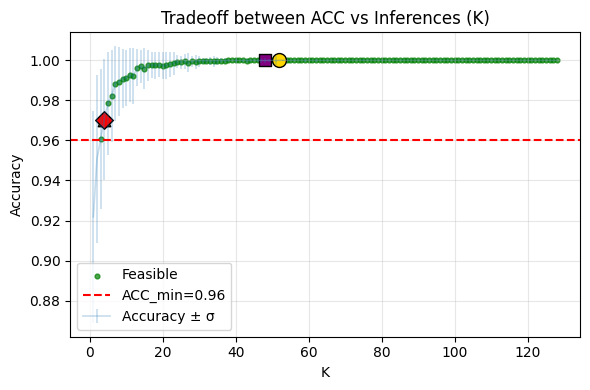}
    \caption{2D ACC--K with optimality for Case~3.}
  \end{subfigure}
    \caption{Accuracy-inference compute (K) trade-offs across three scenarios. These show Monte Carlo means with CI = 95\%; the red dashed line indicates \(A_{\min}\). Distinct operational priorities yield different optimal inference scales \(k^\star\).}
\end{figure}

\subsection{Nvidia Nemotron Nano 9B V2}
This shows the MOO from the Nvidia Nemotron Nano 9B V2 case across three scenarios.

\begin{figure}[H]
  \centering
  \begin{subfigure}{0.32\textwidth}
    \includegraphics[width=\linewidth]{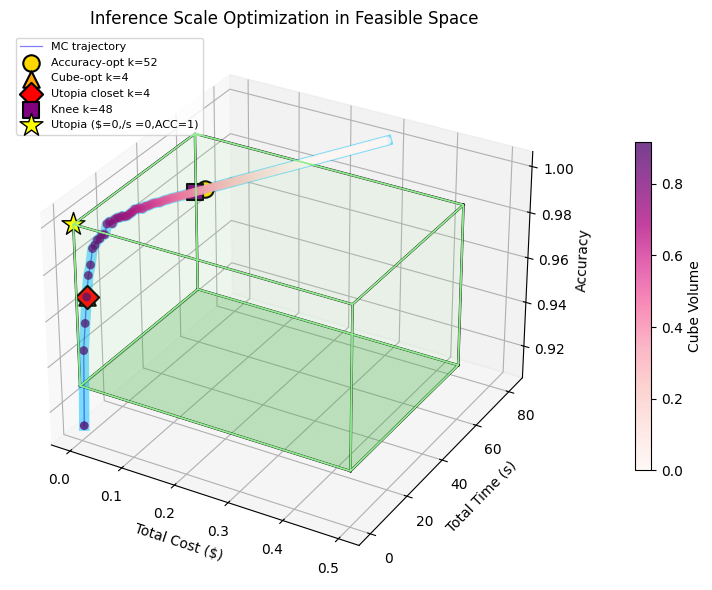}
    \caption{3D MOO results for Case~1.}
  \end{subfigure}
  \begin{subfigure}{0.32\textwidth}
    \includegraphics[width=\linewidth]{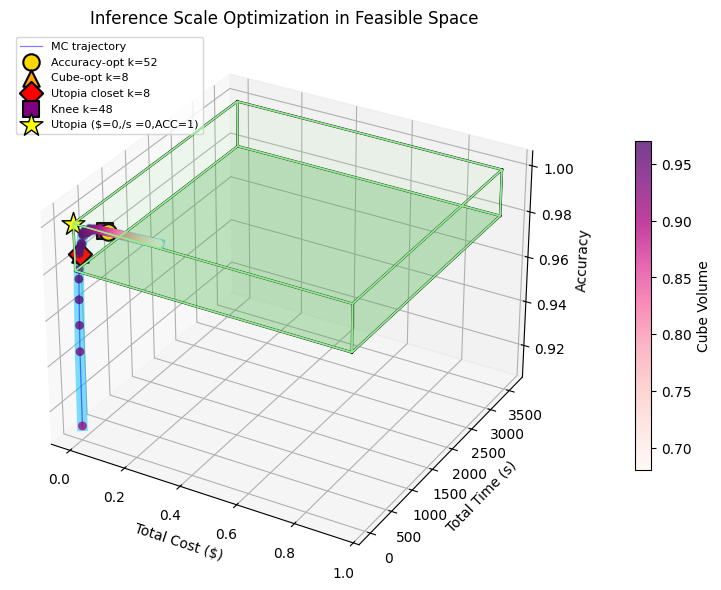}
    \caption{3D MOO results for Case~2.}
  \end{subfigure}
  \begin{subfigure}{0.32\textwidth}
    \includegraphics[width=\linewidth]{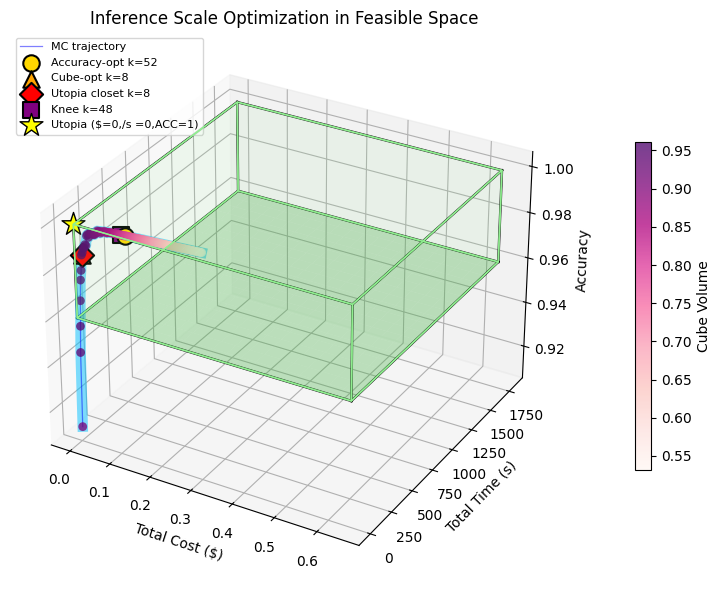}
    \caption{3D MOO results for Case~3.}
  \end{subfigure}
    \caption{
    Inference-scaling optimization results for Nvidia Nemotron Nano 9B V2 across three simulated scenarios. 
    (a--c) 3D feasible cubes in cost--time--accuracy space with constraint planes at \(C_{\max}\), \(T_{\max}\), and \(A_{\min}\). 
    Markers denote optimization criteria (\(\bullet\) Accuracy-Optimal, \(\blacktriangle\) Cube-Optimal, \(\blacklozenge\) Utopia-Closest, \(\blacksquare\) Knee-Point); colors indicate cube volume (larger is better), and the yellow star marks the utopia point.}
\end{figure}

\begin{figure}[H]
  \centering
  \begin{subfigure}{0.32\textwidth}
    \includegraphics[width=\linewidth]{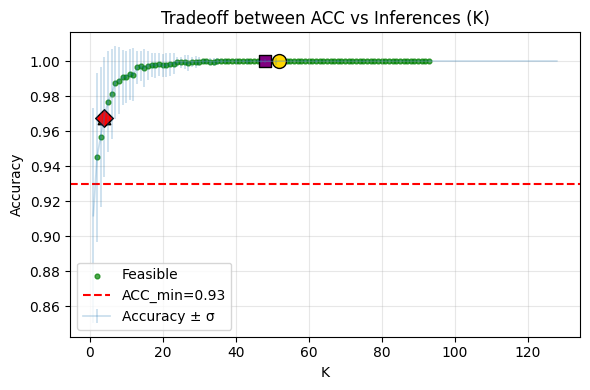}
    \caption{2D ACC--K with optimality for Case~1.}
  \end{subfigure}
  \begin{subfigure}{0.32\textwidth}
    \includegraphics[width=\linewidth]{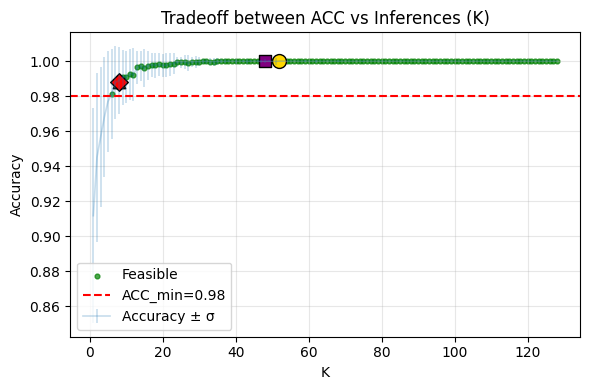}
    \caption{2D ACC--K with optimality for Case~2.}
  \end{subfigure}
  \begin{subfigure}{0.32\textwidth}
    \includegraphics[width=\linewidth]{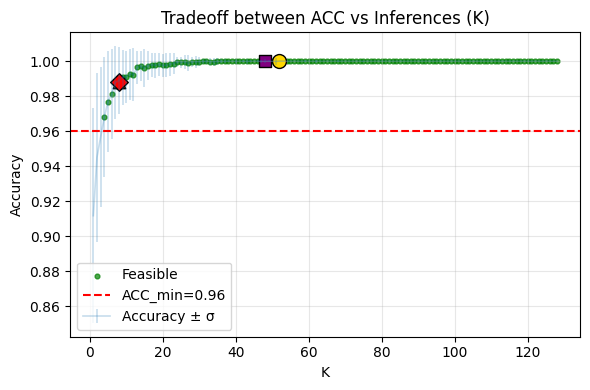}
    \caption{2D ACC--K with optimality for Case~3.}
  \end{subfigure}
    \caption{Accuracy-inference compute (K) trade-offs across three scenarios. These show Monte Carlo means with CI = 95\%; the red dashed line indicates \(A_{\min}\). Distinct operational priorities yield different optimal inference scales \(k^\star\).}
\end{figure}

\subsection{Qwen3-max}
This shows the MOO from the Qwen3-max case across three scenarios.

\begin{figure}[H]
  \centering
  \begin{subfigure}{0.32\textwidth}
    \includegraphics[width=\linewidth]{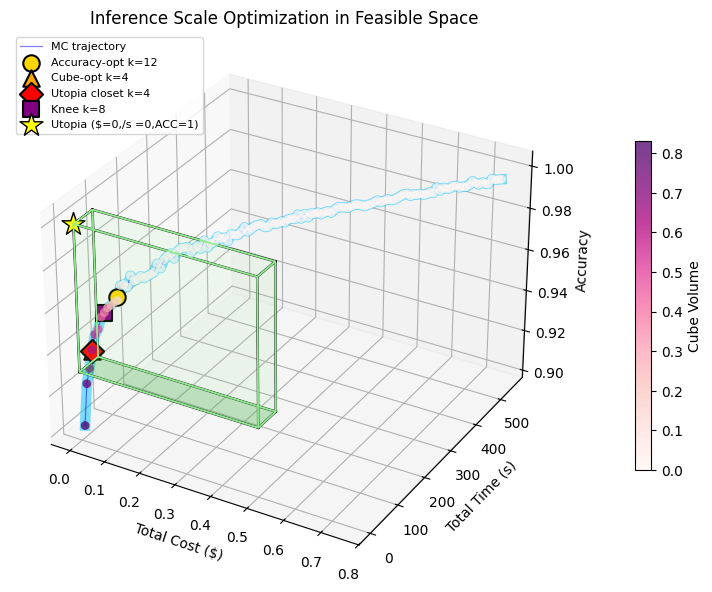}
    \caption{3D MOO results for Case~1.}
  \end{subfigure}
  \begin{subfigure}{0.32\textwidth}
    \includegraphics[width=\linewidth]{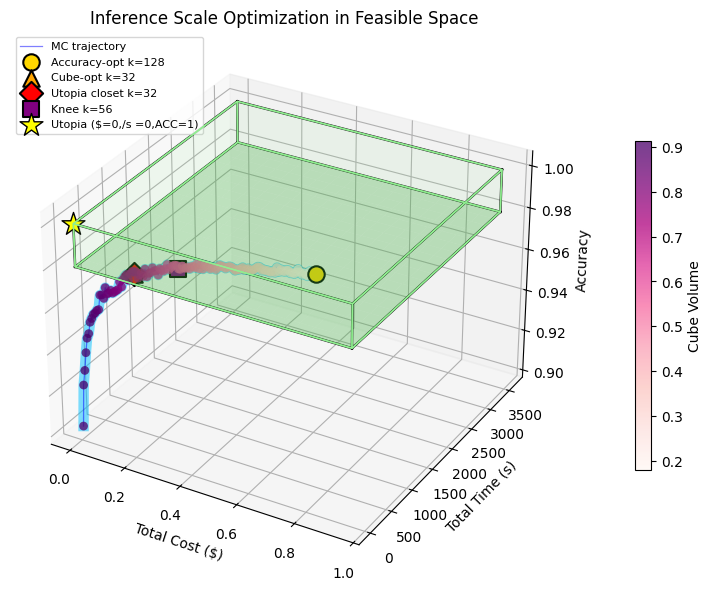}
    \caption{3D MOO results for Case~2.}
  \end{subfigure}
  \begin{subfigure}{0.32\textwidth}
    \includegraphics[width=\linewidth]{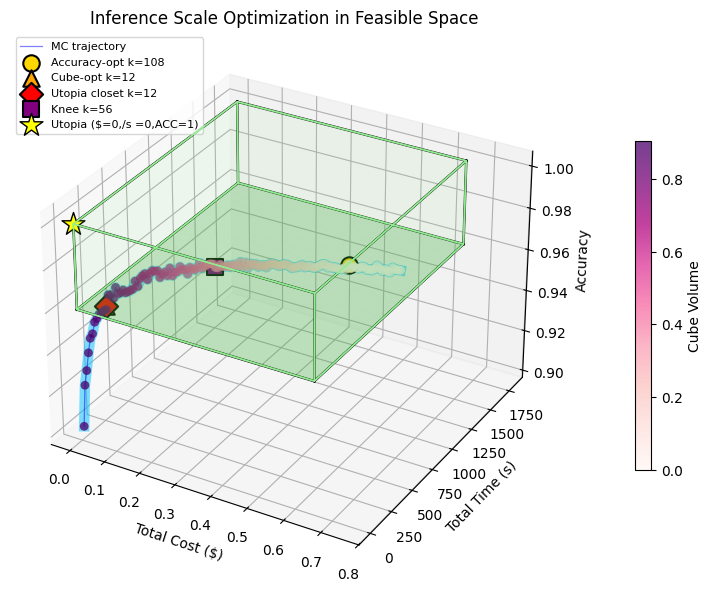}
    \caption{3D MOO results for Case~3.}
  \end{subfigure}
    \caption{
    Inference-scaling optimization results for Qwen3-max across three simulated scenarios. 
    (a--c) 3D feasible cubes in cost--time--accuracy space with constraint planes at \(C_{\max}\), \(T_{\max}\), and \(A_{\min}\). 
    Markers denote optimization criteria (\(\bullet\) Accuracy-Optimal, \(\blacktriangle\) Cube-Optimal, \(\blacklozenge\) Utopia-Closest, \(\blacksquare\) Knee-Point); colors indicate cube volume (larger is better), and the yellow star marks the utopia point.}
\end{figure}

\begin{figure}[H]
  \centering
  \begin{subfigure}{0.32\textwidth}
    \includegraphics[width=\linewidth]{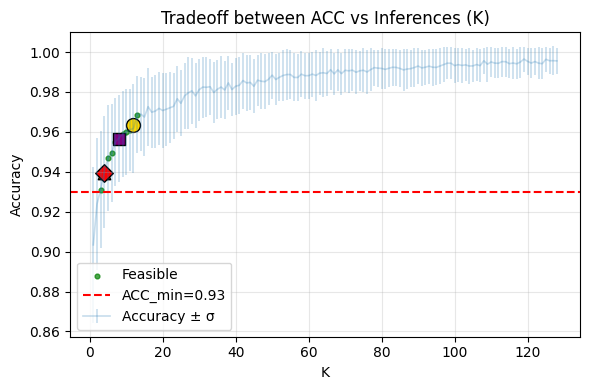}
    \caption{2D ACC--K with optimality for Case~1.}
  \end{subfigure}
  \begin{subfigure}{0.32\textwidth}
    \includegraphics[width=\linewidth]{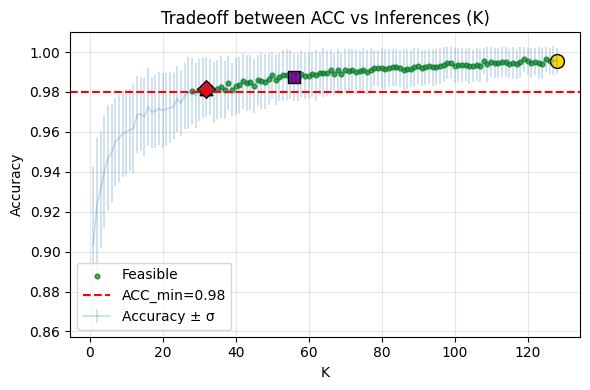}
    \caption{2D ACC--K with optimality for Case~2.}
  \end{subfigure}
  \begin{subfigure}{0.32\textwidth}
    \includegraphics[width=\linewidth]{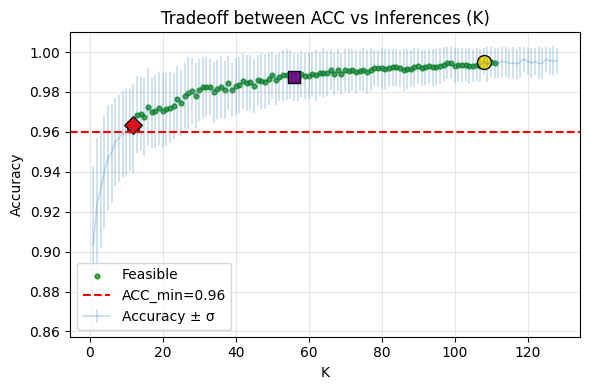}
    \caption{2D ACC--K with optimality for Case~3.}
  \end{subfigure}
    \caption{Accuracy-inference compute (K) trade-offs across three scenarios. These show Monte Carlo means with CI = 95\%; the red dashed line indicates \(A_{\min}\). Distinct operational priorities yield different optimal inference scales \(k^\star\).}
\end{figure}

\subsection{Qwen3-next 30B A3B}
This shows the MOO from the Qwen3-next 80B A3B case across three scenarios.

\begin{figure}[H]
  \centering
  \begin{subfigure}{0.32\textwidth}
    \includegraphics[width=\linewidth]{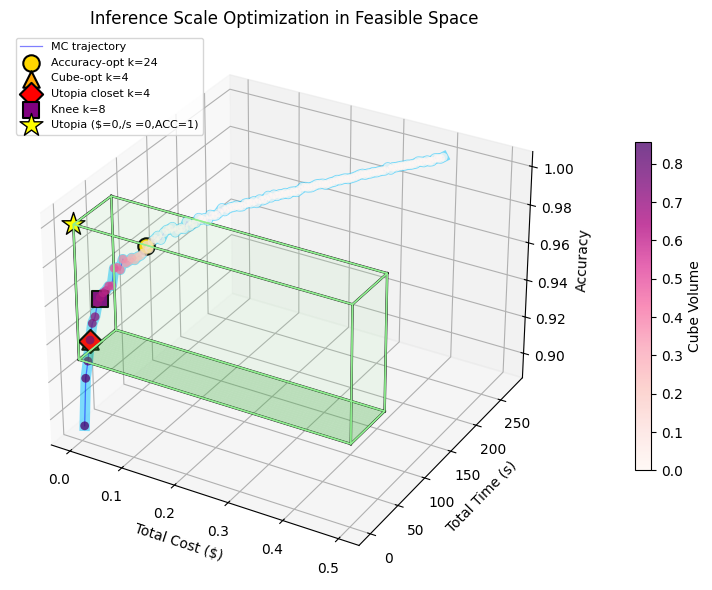}
    \caption{3D MOO results for Case~1.}
  \end{subfigure}
  \begin{subfigure}{0.32\textwidth}
    \includegraphics[width=\linewidth]{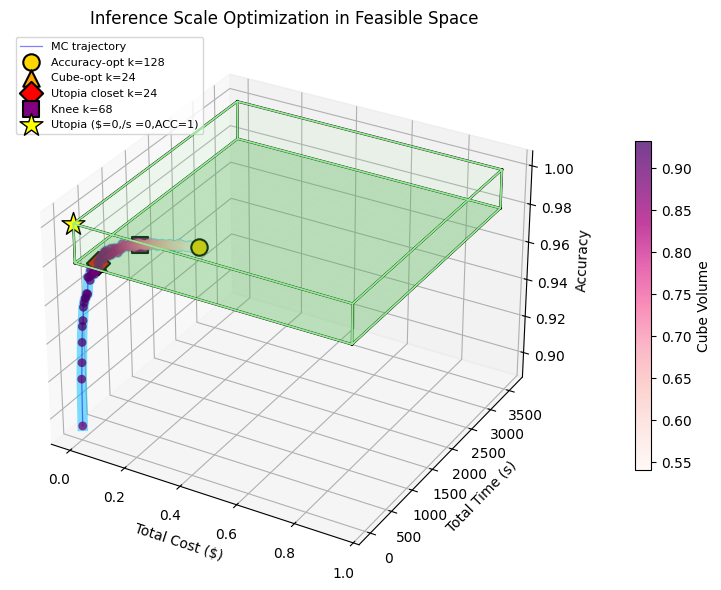}
    \caption{3D MOO results for Case~2.}
  \end{subfigure}
  \begin{subfigure}{0.32\textwidth}
    \includegraphics[width=\linewidth]{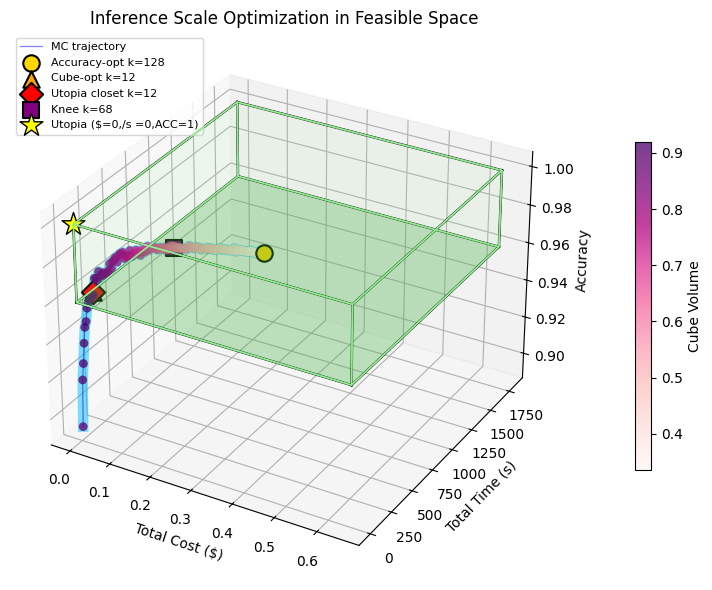}
    \caption{3D MOO results for Case~3.}
  \end{subfigure}
    \caption{
    Inference-scaling optimization results for Qwen3-next 80B A3B across three simulated scenarios. 
    (a--c) 3D feasible cubes in cost--time--accuracy space with constraint planes at \(C_{\max}\), \(T_{\max}\), and \(A_{\min}\). 
    Markers denote optimization criteria (\(\bullet\) Accuracy-Optimal, \(\blacktriangle\) Cube-Optimal, \(\blacklozenge\) Utopia-Closest, \(\blacksquare\) Knee-Point); colors indicate cube volume (larger is better), and the yellow star marks the utopia point.}
\end{figure}

\begin{figure}[H]
  \centering
  \begin{subfigure}{0.32\textwidth}
    \includegraphics[width=\linewidth]{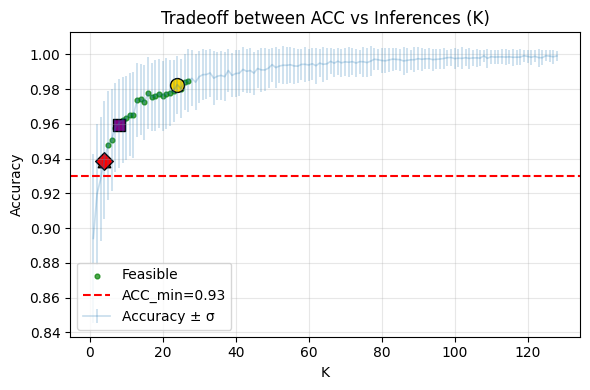}
    \caption{2D ACC--K with optimality for Case~1.}
  \end{subfigure}
  \begin{subfigure}{0.32\textwidth}
    \includegraphics[width=\linewidth]{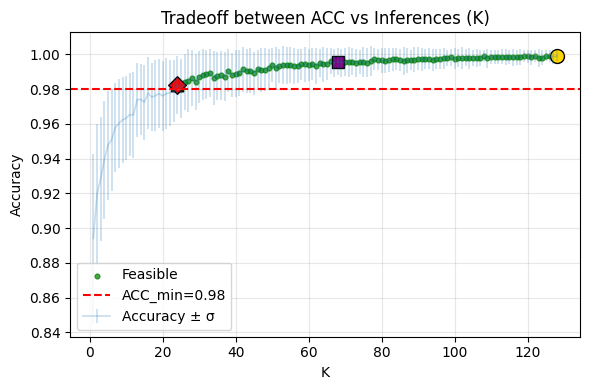}
    \caption{2D ACC--K with optimality for Case~2.}
  \end{subfigure}
  \begin{subfigure}{0.32\textwidth}
    \includegraphics[width=\linewidth]{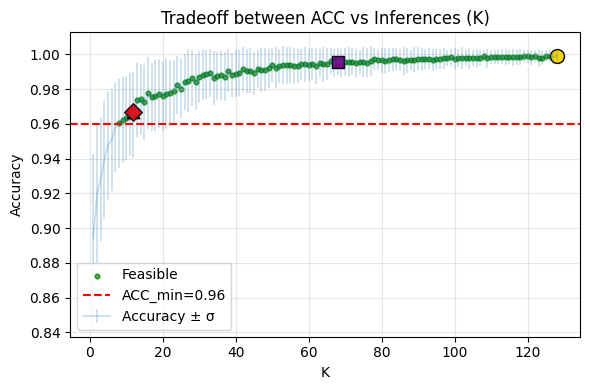}
    \caption{2D ACC--K with optimality for Case~3.}
  \end{subfigure}
    \caption{Accuracy-inference compute (K) trade-offs across three scenarios. These show Monte Carlo means with CI = 95\%; the red dashed line indicates \(A_{\min}\). Distinct operational priorities yield different optimal inference scales \(k^\star\).}
\end{figure}

\subsection{Qwen3 30B A3B}
This shows the MOO from the Qwen3 30B A3B case across three scenarios.

\begin{figure}[H]
  \centering
  \begin{subfigure}{0.32\textwidth}
    \includegraphics[width=\linewidth]{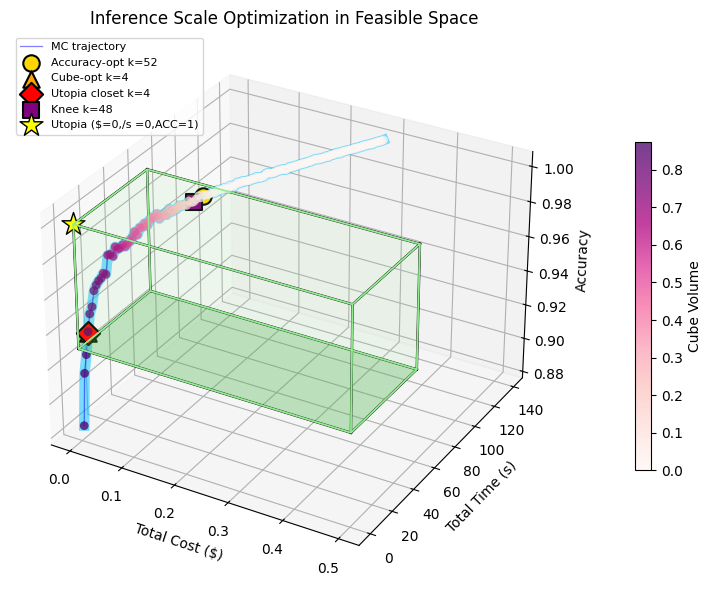}
    \caption{3D MOO results for Case~1.}
  \end{subfigure}
  \begin{subfigure}{0.32\textwidth}
    \includegraphics[width=\linewidth]{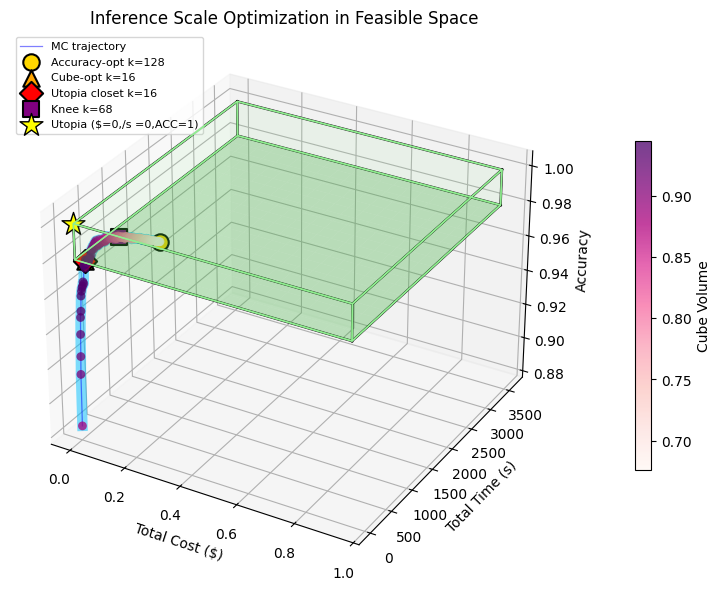}
    \caption{3D MOO results for Case~2.}
  \end{subfigure}
  \begin{subfigure}{0.32\textwidth}
    \includegraphics[width=\linewidth]{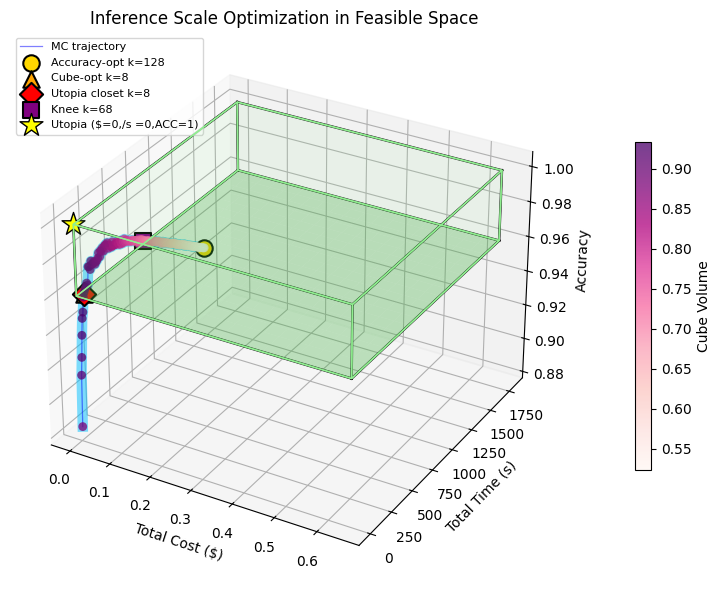}
    \caption{3D MOO results for Case~3.}
  \end{subfigure}
    \caption{
    Inference-scaling optimization results for Qwen3 30B A3B across three simulated scenarios. 
    (a--c) 3D feasible cubes in cost--time--accuracy space with constraint planes at \(C_{\max}\), \(T_{\max}\), and \(A_{\min}\). 
    Markers denote optimization criteria (\(\bullet\) Accuracy-Optimal, \(\blacktriangle\) Cube-Optimal, \(\blacklozenge\) Utopia-Closest, \(\blacksquare\) Knee-Point); colors indicate cube volume (larger is better), and the yellow star marks the utopia point.}
\end{figure}

\begin{figure}[H]
  \centering
  \begin{subfigure}{0.32\textwidth}
    \includegraphics[width=\linewidth]{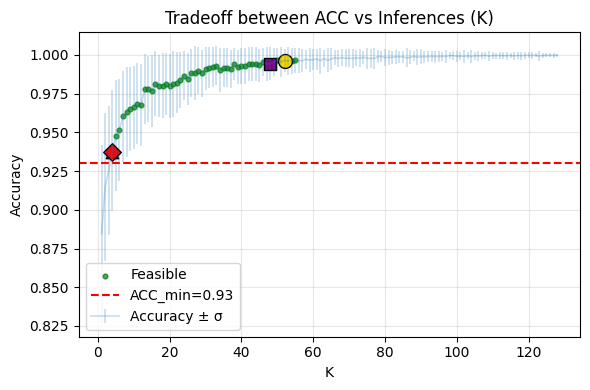}
    \caption{2D ACC--K with optimality for Case~1.}
  \end{subfigure}
  \begin{subfigure}{0.32\textwidth}
    \includegraphics[width=\linewidth]{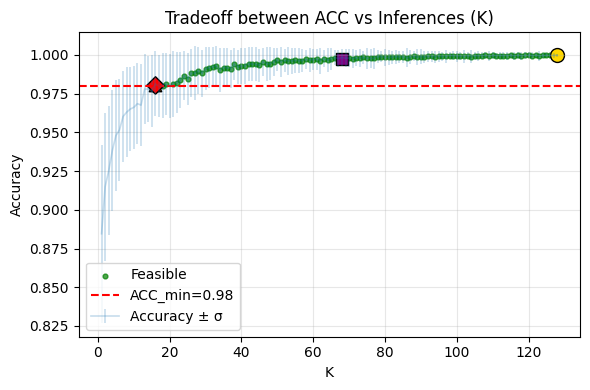}
    \caption{2D ACC--K with optimality for Case~2.}
  \end{subfigure}
  \begin{subfigure}{0.32\textwidth}
    \includegraphics[width=\linewidth]{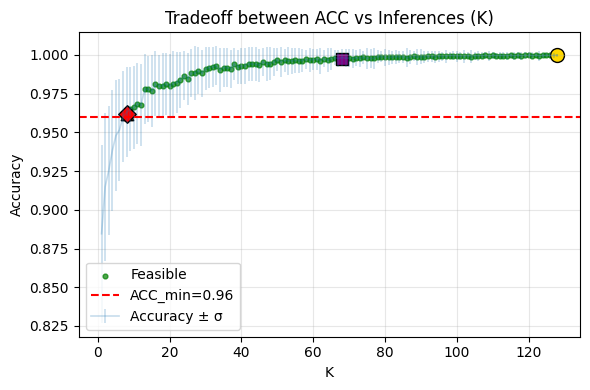}
    \caption{2D ACC--K with optimality for Case~3.}
  \end{subfigure}
    \caption{Accuracy-inference compute (K) trade-offs across three scenarios. These show Monte Carlo means with CI = 95\%; the red dashed line indicates \(A_{\min}\). Distinct operational priorities yield different optimal inference scales \(k^\star\).}
\end{figure}

\end{document}